\documentclass[final,3p,twocolumn]{elsarticle}

\usepackage{mathptmx}           

\usepackage[protrusion=true]{microtype}

\usepackage{amsmath,amssymb}    
\usepackage{graphicx}
\graphicspath{{paper_figures/}}
\usepackage{booktabs}
\usepackage{longtable}
\usepackage{array}
\usepackage{tabularx}
\usepackage{multirow}
\usepackage{textcomp}
\usepackage{float}
\usepackage{enumitem}
\usepackage{algorithm}
\usepackage{algpseudocode}
\usepackage{xcolor}
\usepackage{tikz}
\usetikzlibrary{arrows.meta,shapes.geometric,positioning,fit,backgrounds,calc,decorations.pathmorphing}

\usepackage{hyperref}
\usepackage{xurl}
\hypersetup{
  colorlinks = false,    
  hidelinks,
  pdfauthor  = {Puneet Kant},
  pdftitle   = {A Synthetic Reliability-Aware PINN Benchmark for Offshore Wind Turbine Support-Structure Monitoring with Bayesian Inverse Identification}
}

\makeatletter
\def\printFirstPageNotes{%
  \iflongmktitle
    \let\columnwidth=\textwidth
  \fi
  \ifdoubleblind
  \else
    \ifx\@tnotes\@empty\else\@tnotes\fi
    \ifx\@nonumnotes\@empty\else\@nonumnotes\fi
    \ifx\@cornotes\@empty\else\@cornotes\fi
    \let\thefootnote\relax
    \footnotetext{%
      \noindent
      \begin{tabular}[t]{@{}l@{}}
        \textit{Email addresses:} \texttt{m25ai1078@iitj.ac.in} (Puneet Kant),\\
        \texttt{monikatanwar@iitj.ac.in} (Monika Tanwar)
      \end{tabular}}%
    \ifx\@elsuads\@empty\relax\else
      \let\thefootnote\relax
      \footnotetext{\textit{URL:\space}\@elsuads}%
    \fi
  \fi
  \ifx\@fnotes\@empty\else\@fnotes\fi
  \iflongmktitle\if@twocolumn
    \let\columnwidth=\Columnwidth\fi\fi
}
\makeatother

\begin{document}

\begin{frontmatter}

\title{A Synthetic Reliability-Aware PINN Benchmark for Offshore Wind Turbine
Support-Structure Monitoring with Bayesian Inverse Identification}

\author[saids]{Puneet Kant\corref{cor1}\fnref{fn1}}
\ead{m25ai1078@iitj.ac.in}

\author[sme]{Monika Tanwar}
\ead{monikatanwar@iitj.ac.in}

\address[saids]{School of Artificial Intelligence and Data Science,
Indian Institute of Technology Jodhpur}
\address[sme]{School of Management and Entrepreneurship,
Indian Institute of Technology Jodhpur}

\cortext[cor1]{Corresponding author.}
\fntext[fn1]{Manuscript prepared June 2026.}


\begin{abstract}
Reliable structural health monitoring (SHM) of offshore wind turbine (OWT)
support structures requires fast state estimation from sparse measurements.
Repeated high-fidelity finite-element or aeroelastic analyses are difficult to
use directly in online monitoring loops, while purely data-driven surrogates can
require large training sets. This paper presents \textbf{DigiTurbine}, a
synthetic reliability-aware Physics-Informed Neural Network (PINN) benchmark for
OWT monopile support-structure monitoring. The workflow embeds a simplified
Euler--Bernoulli beam equation with Winkler soil foundation in the training
objective, couples it with Bayesian-prior-informed inverse identification, and
adds First-Order Reliability Method (FORM) screening. All validation uses
synthetic configurations with analytical or finite-difference ground truth
motivated by the NREL 5\,MW reference turbine context.

Three contributions are evaluated.
\textbf{(1) Forward PINN:} A 4,353-parameter network with EMA-adaptive loss
weighting achieves mean RMSE $0.135 \pm 0.109$\,mm (10/10 pass) with
all-configuration mean inference of 0.381\,ms (GPU) / 0.605\,ms (CPU),
26$\times$ / 17$\times$ below the 10\,ms real-time target.
\textbf{(2) Inverse PINN and gradient direction problem:} Standard simultaneous
optimisation of network weights and unknown parameters ($E$,
$k_\text{soil}$) fails on the tested 4th-order PDE inverse cases because of
conflicting gradient directions (mean error 63.9\,\%, 0/4 pass), and five
loss-weighting or scheduling mitigations also fail in the large-offset
synthetic tests. A weak log-normal Bayesian prior centred on the correct
design/as-built specification substantially mitigates this failure on the
tested cases: 8/8 pass, mean $E$ error $\leq 0.02$\,\%, including simultaneous
$E{+}k_\text{soil}$ identification under 10\,\% measurement noise.
\textbf{(3) Near-real-time reliability:} The FORM layer solves representative
structural limit-state cases in 0.7--2.7\,ms, with the OWT cantilever
root-moment capacity case requiring about 1.0\,ms ($>$40$\times$ faster than
Monte Carlo), machine-precision agreement for linear limit states, and
$\leq 6$\,\% error for moderately nonlinear cases. In the benchmarked synthetic
workflow, the online pipeline completes in $<$7\,ms, giving theoretical
processing headroom for SCADA-style monitoring.
\end{abstract}

\begin{keyword}
Physics-Informed Neural Networks \sep Synthetic Benchmark \sep Digital Twin \sep
Offshore Wind Turbine \sep Structural Health Monitoring \sep
Bayesian Inverse Identification \sep Gradient Direction Problem \sep
First-Order Reliability Method \sep Euler--Bernoulli Beam
\end{keyword}

\end{frontmatter}


\section{Introduction}
\label{sec:intro}

Global wind installations reached a record 165\,GW in 2025, taking total
installed wind capacity to 1{,}299\,GW, while offshore wind reached 92.5\,GW by
the end of 2025 after 9.3\,GW of new offshore capacity was grid-connected
\citep{GWECGlobalWind2026,GWECOffshore2026}. Continued renewable expansion is
expected toward 2030 \citep{IEA2025}. Fixed-bottom support structures, especially
monopiles, remain central to deployed offshore wind and must endure
$10^8$--$10^9$ fatigue cycles over a 25-year service life, making fatigue a
dominant structural design concern
\citep{DNV2021,Augustyn2021,Wang2021}. Operations and maintenance (O\&M) costs
represent 25--30\,\% of lifecycle expenditure, and published condition-based
monitoring studies report potential O\&M savings of 11--18\,\%
\citep{LeonMedina2025}. Yet structural sensing remains sparse, often limited to
small sets of accelerometer and strain measurements \citep{Augustyn2021,Wang2022,LeonMedina2025},
making real-time full-field structural state inference from sparse, noisy data a
fundamental challenge.

Physics-Informed Neural Networks (PINNs) \citep{Raissi2019} encode governing
PDEs as soft loss constraints, enabling sparse-data surrogates in solid mechanics
and wind-energy applications \citep{Haghighat2021,Chen2025}. The present study
tests the 50--100 measurement regime, far fewer than would typically be expected
for purely data-driven surrogates. Extended to inverse problems, PINNs treat
unknown material properties as additional optimisation variables; however, a
\textit{gradient direction problem}, not systematically characterised
in the reviewed literature, prevents convergence in the tested cases when
parameters are initialised far from their true values. Digital Twins
(DTs) at Level~2--3 maturity \citep{Rasheed2020,Ritto2021} couple physics-based
surrogates with reliability analysis for continuous structural prognostics.
First-Order Reliability Method (FORM) \citep{Wang2022} estimates the reliability
index $\beta(t)$ through constrained optimisation and is typically far cheaper
than Monte Carlo for low-dimensional limit states; paired with PINN-updated
structural estimates, it enables near-real-time reliability screening not found
in the reviewed OWT DT literature.

\subsection{Research Gaps}
\label{sec:intro_gaps}

A focused review of recent PINN, digital-twin, reliability, and OWT structural
monitoring studies identifies four open gaps; representative studies are shown
in Table~\ref{tab:litgap}.

\begin{enumerate}[label=\textbf{G\arabic*.}]
  \item \textbf{Computational gap.} High-fidelity FEA can require minutes to hours per evaluation; pure ML requires
        thousands of training samples. No OWT SHM framework in the reviewed literature was found to achieve sub-10\,ms
        physics-consistent inference from sparse data.

  \item \textbf{Inverse problem gap.} Published inverse PINNs do not study convergence
        failure under large-offset initialisations. The gradient direction
    problem was not found to be identified, formally analysed, or substantially mitigated in the reviewed literature.

        \item \textbf{Integration gap.} Within the reviewed literature, no prior work was found to integrate PINN forward prediction,
          inverse parameter identification, and probabilistic reliability in a single
          synthetic-data-validated OWT pipeline.

    \item \textbf{Reliability speed gap.} Monte Carlo can be too slow for repeated
      online reliability updates, particularly when coupled to high-fidelity
      structural models; FORM integrated with a physics-consistent PINN
      state-estimation workflow was not found for OWT structures in the reviewed
      literature.

\end{enumerate}

\subsection{Contributions}
\label{sec:intro_contributions}

This paper makes the following specific contributions, evaluated using synthetic benchmark experiments.

\begin{enumerate}
  \item \textbf{Forward PINN for OWT monopile:} PDE-constrained surrogate for a
        simplified Euler--Bernoulli/Winkler monopile benchmark (a Morison
        forcing helper is implemented in \texttt{morison\_force()}, but
        Morison-driven PINN training is deferred to future field-calibration work),
        evaluated across 10 synthetic uniform-load configurations.
        Mean RMSE $0.135 \pm 0.109$\,mm,
        all-configuration mean inference 0.381\,ms (GPU) / 0.605\,ms (CPU)
        (both well below the 10\,ms target);
        10/10 test cases pass from only 50--100 sparse measurements.

  \item \textbf{Gradient direction problem: identification, analysis, and Bayesian
        mitigation.} Formal gradient analysis and experimental characterisation of the
        circular-dependency failure in standard inverse PINNs for high-order PDEs.
        Five mitigation strategies all fail without priors. Weak Bayesian priors
        centred on the correct design/as-built specifications achieve 8/8 success, mean $E$ error
        $\leq 0.02$\,\%, including simultaneous $E + k_\text{soil}$ identification
        under 10\,\% noise.

  \item \textbf{FORM reliability layer within the PINN workflow:} representative
        FORM solves in 0.7--2.7\,ms, with the OWT cantilever root-moment capacity case
        at about 1.0\,ms
        ($>40\times$ faster than Monte Carlo); machine-precision accuracy for linear
        limit states; ${\leq}6\,\%$ error for moderately nonlinear cases.
        DigiTurbine end-to-end pipeline: $<$7\,ms, corresponding to
        $>$143\,Hz theoretical processing headroom.

\end{enumerate}

\section{Literature Review}
\label{sec:litreview}

\subsection{Digital Twins and PINNs for OWT}
\label{sec:lit_dt}

Digital twin applications to OWT have grown rapidly, covering fleet-level SCADA
analytics \citep{Ambarita2024}, component-level SHM \citep{LeonMedina2025},
Bayesian reliability updating from monitoring data \citep{Augustyn2021},
probabilistic risk-based inspection planning \citep{Bull2024}, and mooring-line
monitoring \citep{Walker2022}. Wang et~al.\ \citep{Wang2021} conducted
a comprehensive review of OWT support-structure reliability within a DT
context, concluding that no ML- or PINN-based framework existed for integrated
reliability monitoring. Chiach\'io et~al.\ \citep{Chiachio2022} and
Ritto \& Rochinha \citep{Ritto2021} developed DT frameworks with Bayesian updating
and physics-based ML classifiers respectively, but without PINN-based millisecond
inference. Chen et~al.\ \citep{Chen2024} reviewed floating-OWT DTs using hybrid
neural surrogates; Lai et~al.\ \citep{Lai2023} demonstrated fatigue monitoring
in aircraft wings via measurement--computation fusion; both lack real-time
reliability metrics.

PINNs \citep{Raissi2019,Karniadakis2021} have been applied to solid-mechanics
material identification \citep{Haghighat2021}, wind-farm wake dynamics \citep{Zhang2023},
wind-turbine bearing fatigue \citep{Yucesan2023}, wind-turbine power prediction
with uncertainty quantification \citep{Gijon2023}, and monopile scour detection
from natural-frequency shifts \citep{Chen2025}. These works collectively
establish PINNs as viable for wind-energy applications, but the reviewed works
do not address convergence failure of inverse PINNs or integrate structural
reliability analysis.

\subsection{Inverse PINNs and Structural Reliability}
\label{sec:lit_inverse}

Standard inverse PINNs co-optimise network weights and unknown physical parameters
via a PDE-constrained loss \citep{Raissi2019}. Applications to bearing diagnostics
\citep{Qin2024} and elasticity identification \citep{Haghighat2021} report success
but generally do not systematically stress-test large-offset initialisations.
Offshore SHM can involve substantial parameter uncertainty from material
degradation, scour, and soil--structure changes; convergence under such
conditions remains underexplored. Published works report successful cases without
examining failure mechanisms, and no systematic study of convergence under varied
initialisations or noise levels was found. In particular, the gradient direction
problem (large-error failure from circular dependencies between unknown
parameters and high-order network derivatives) has not previously been
identified, characterised, or substantially mitigated in the reviewed literature.

Structural reliability for OWT has been addressed via FORM, SORM, and Monte
Carlo \citep{Wang2022,Das2025}, but probabilistic analysis remains computationally
intractable for continuous monitoring at FEA speeds. FORM integrated with a
physics-consistent PINN state-estimation workflow for real-time $\beta(t)$ was
not found in the reviewed literature for OWT support structures.

\subsection{Gap Matrix and Positioning}
\label{sec:lit_gap}

Table~\ref{tab:litgap} presents a positioning matrix across the reviewed
literature. The bottom row summarises the limited-scope combination implemented
in this synthetic benchmark.

\begin{table*}[htbp]
\caption{Literature positioning matrix across the five target capability
dimensions. Symbols denote coverage within each work's stated validation scope;
descriptive entries specify the narrower implementation scope when a simple
checkmark would be ambiguous. \checkmark~=~substantive coverage;
$\sim$~=~partial coverage; --~=~not addressed.}
\label{tab:litgap}
\centering
\small
\setlength{\tabcolsep}{3pt}
\begin{tabular}{llccccc}
\toprule
\textbf{Reference} & \textbf{Venue / Year} & \textbf{DT} & \textbf{PINN} &
\textbf{Reliability} & \textbf{OWT Struct.} & \textbf{Inverse} \\
\midrule
Leon-Medina et~al.\ \citep{LeonMedina2025} & Intell.~Syst.~Appl.\ 2025
  & \checkmark & ML only & Component & \checkmark & -- \\
Ambarita et~al.\ \citep{Ambarita2024} & Energy Inform.\ 2024
  & \checkmark & -- & -- & \checkmark & -- \\
Chen et~al.\ \citep{Chen2024} & Energies 2024
  & $\sim$ & Hybrid & Qualitative & Floating & -- \\
Augustyn et~al.\ \citep{Augustyn2021} & Energies 2021
  & \checkmark & -- & Bayesian & \checkmark & -- \\
Bull et~al.\ \citep{Bull2024} & Struct.~Health Monit.\ 2025
  & \checkmark & -- & Risk-based & \checkmark & -- \\
Walker et~al.\ \citep{Walker2022} & J.~Ocean~Eng.\ 2022
  & \checkmark & -- & Component & Mooring & -- \\
Qin et~al.\ \citep{Qin2024} & Knowl.-Based Syst.\ 2024
  & \checkmark & Inverse & Diagnostics & -- & \checkmark \\
Zhang \& Zhao \citep{Zhang2023} & En.~Conv.~Mgmt.\ 2023
  & \checkmark & Farm & -- & \checkmark & -- \\
Chen et~al.\ \citep{Chen2025} & Ocean Eng.\ 2025
  & -- & Forward & -- & Monopile & $\sim$ \\
Das et~al.\ \citep{Das2025} & Comput.~Struct.\ 2025
  & -- & Bayesian & General & -- & -- \\
Chiachio et~al.\ \citep{Chiachio2022} & Autom.~Constr.\ 2022
  & \checkmark & -- & $\sim$ & -- & -- \\
Ritto \& Rochinha \citep{Ritto2021} & Mech.~Syst.~Signal 2021
  & \checkmark & ML class. & -- & \checkmark & -- \\
Lai et~al.\ \citep{Lai2023} & J.~Manuf.~Syst.\ 2023
  & \checkmark & -- & Fatigue & -- & -- \\
Wang et~al.\ \citep{Wang2021} & Ocean Eng.\ 2021
  & $\sim$ & -- & Review & \checkmark & -- \\
Yucesan \& Viana \citep{Yucesan2023} & Appl.~Soft~Comput.\ 2023
  & \checkmark & \checkmark & Bearing & Component & -- \\
\midrule
\textbf{This work} & Synthetic benchmark
  & Synth. DT & Fwd.+inv. PINN & FORM & 1D monopile & Bayes prior \\
\addlinespace[2pt]
\multicolumn{7}{@{}p{0.98\textwidth}@{}}{\footnotesize\textit{Note:} For this work,
the bottom row denotes implemented components in the present synthetic
benchmark, not field deployment: reliability uses algebraic FORM screening, OWT
structural modelling uses a one-dimensional monopile beam, and inverse
identification uses a correctly centred weak prior; see
Table~\ref{tab:scope_assumptions}.} \\
\bottomrule
\end{tabular}
\end{table*}

\section{Theoretical Framework}
\label{sec:theory}

\subsection{Structural Mechanics of the OWT Monopile}
\label{sec:theory_struct}

\subsubsection{Euler--Bernoulli Beam with Winkler Foundation}
\label{sec:theory_eb}

The transverse vibration of a monopile section is governed by the
Euler--Bernoulli partial differential equation in the presence of viscous damping,
distributed loading, and Winkler soil reaction:

\begin{equation}
\begin{split}
EI\,\frac{\partial^4 u}{\partial x^4}
  &+ k_{\text{soil}}(x)\,u
   + c\,\frac{\partial u}{\partial t} \\
  &+ \rho A\,\frac{\partial^2 u}{\partial t^2}
   = q(x,t),
\end{split}
\label{eq:EB}
\end{equation}

where $x \in [0,L]$ is the axial coordinate, $t$ is time, $u(x,t)$ is the
transverse displacement (m), $EI$ is the bending stiffness (Pa\,m$^4$), $\rho A$
is the linear mass (kg\,m$^{-1}$), $c$ is the distributed damping coefficient
(N\,s\,m$^{-2}$), $k_\text{soil}$ is the equivalent Winkler foundation stiffness
per unit beam length used in the one-dimensional beam equation (N\,m$^{-2}$),
and $q(x,t)$ is the distributed transverse load (N\,m$^{-1}$).

For the NREL 5\,MW reference turbine \citep{jonkman2009definition} the relevant
nominal properties are:
pile outer diameter $D = 6.0$\,m, wall thickness $t_w = 0.06$\,m, Young's modulus
$E = 210$\,GPa, density $\rho = 7{,}850$\,kg\,m$^{-3}$, cross-sectional area
$A = 1.13$\,m$^2$, second moment of area $I = 4.94$\,m$^4$, and pile embedment
depth 30\,m. These properties are cited for physical context of the NREL 5\,MW
reference turbine; the validation experiments in Section~\ref{sec:results} use a
representative simplified beam model ($I = 8.333 \times 10^{-3}$\,m$^4$, $A = 0.05$\,m$^2$,
$EI = 1.75$\,GN\,m$^2$) that produces well-resolved millimetre-scale deflections,
enabling exact ground-truth comparison with the analytical solution
(Eq.~\ref{eq:analytical}).
Representative geotechnical subgrade moduli range from 5\,MN\,m$^{-3}$ (loose
sand) to 60\,MN\,m$^{-3}$ (dense sand or stiff clay) according to DNV-ST-0126
\citep{DNV2021,Wang2022}. When used in Eq.~\eqref{eq:EB}, these values are
converted to an equivalent per-length beam stiffness by multiplying by an
effective pile width or diameter.

\textbf{Boundary conditions} for a clamped-base (pile--soil interface), free-top
cantilever:
\begin{equation}
\begin{gathered}
u(0,t) = 0, \quad u_x(0,t) = 0, \\
EI\,u_{xx}(L,t) = 0, \quad EI\,u_{xxx}(L,t) = 0.
\end{gathered}
\label{eq:BC}
\end{equation}

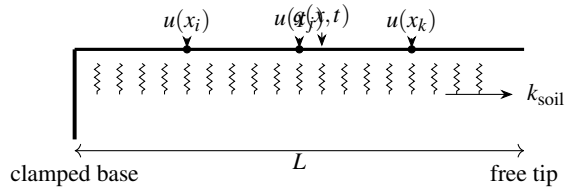
\begin{figure}[!t]
\centering
\resizebox{0.95\columnwidth}{!}{%
\begin{tikzpicture}[scale=0.75, every node/.style={font=\small, inner sep=1pt}]
  \draw[line width=1.2pt] (0,0) -- (8,0);
  \draw[line width=1.5pt] (0,0) -- (0,-1.6);
  \foreach \x in {0.4,0.8,...,7.6} {
    \draw[decorate,decoration={zigzag,segment length=3pt,amplitude=1pt}] (\x,-0.25) -- +(0,-0.55);
  }
  \draw[->,>=Stealth] (4.4,0.3) -- (4.4,0.05);
  \node[above] at (4.4,0.33) {$q(x,t)$};
  \foreach \x in {2.0,4.0,6.0} {
    \draw[fill=black] (\x,0) circle (1.8pt);
    \draw[->,>=Stealth] (\x,0.2) -- (\x,0.02);
  }
  \node[above] at (2.0,0.24) {$u(x_i)$};
  \node[above] at (4.0,0.24) {$u(x_j)$};
  \node[above] at (6.0,0.24) {$u(x_k)$};
  \draw[<->] (0,-1.8) -- node[below] {$L$} (8,-1.8);
  \draw[->,>=Stealth] (6.6,-0.8) -- (7.8,-0.8);
  \node[right] at (8.0,-0.8) {$k_{\text{soil}}$};
  \node[below] at (0,-2.0) {clamped base};
  \node[below] at (8,-2.0) {free tip};
\end{tikzpicture}%
}
\caption{Simplified monopile beam model and measurement layout used in this paper.}
\label{fig:monopile_model}
\end{figure}

Figure~\ref{fig:monopile_model} illustrates the simplified monopile model used in
this study, showing the clamped base, soil reaction, distributed loading, and
sparse displacement measurement locations.

The analytical static solution for a uniformly distributed load $q$ on a
clamped-free beam without soil interaction (used for validation):
\begin{equation}
\begin{split}
u(x) &= \frac{q x^2}{24EI}\bigl(6L^2 - 4Lx + x^2\bigr).
\end{split}
\label{eq:analytical}
\end{equation}
This expression provides exact ground truth for the no-soil static beam cases;
finite-difference Winkler-beam references are used for the soil-interaction
inverse cases.

\subsubsection{Morison Hydrodynamic Loading}
\label{sec:theory_morison}

Wave loading on the sub-surface monopile section can be represented by the
Morison equation \citep{Morison1950}:
\begin{equation}
q_{\text{wave}}(x,t)
  = \tfrac{1}{2}\rho_w C_D D|v_w|v_w
  + \rho_w C_M \tfrac{\pi D^2}{4}\dot{v}_w,
\label{eq:morison}
\end{equation}
where $\rho_w = 1{,}025$\,kg\,m$^{-3}$, $C_D \in [0.6,1.2]$, and
$C_M \in [1.2,2.0]$. The current implementation
(\texttt{morison\_force()}) is a simplified Morison forcing helper used for
software validation; Airy/JONSWAP wave-kinematics coupling and full aero-elastic
Morison-driven training are deferred to future field-calibration work.

\subsection{PINN Formulation}
\label{sec:theory_pinn}

\subsubsection{Composite Loss Function}
\label{sec:theory_loss}

A PINN approximates $u(x,t)$ with a multilayer perceptron
$\hat{u}(x,t;\theta)$ parameterised by trainable weights~$\theta$.
The training objective is the weighted composite loss:

\begin{equation}
\mathcal{L}(\theta) =
  \lambda_{\text{data}}\,\mathcal{L}_{\text{data}}(\theta)
+ \lambda_{\text{PDE}}\,\mathcal{L}_{\text{PDE}}(\theta)
+ \lambda_{\text{BC}}\,\mathcal{L}_{\text{BC}}(\theta),
\label{eq:losstotal}
\end{equation}

where the three loss components are defined as follows.

\textbf{Data loss} (fit to sparse sensor observations):
\begin{equation}
\mathcal{L}_{\text{data}}(\theta)
= \frac{1}{N_d}\sum_{i=1}^{N_d}
  \bigl[\hat{u}(x_i,t_i;\theta) - u_i^{\text{obs}}\bigr]^2.
  \label{eq:Ldata}
\end{equation}

\textbf{PDE residual loss} (enforce Euler--Bernoulli equation at $N_f$
collocation points uniformly distributed throughout the domain):
\begin{equation}
\begin{split}
\mathcal{L}_{\text{PDE}}(\theta)
  &= \frac{1}{N_f}\sum_{j=1}^{N_f}
  \Bigl[EI\,\hat{u}_{xxxx,j}
       + k_\text{soil}\,\hat{u}_j \\
  &\quad + c\,\dot{\hat{u}}_j
       + \rho A\,\ddot{\hat{u}}_j
       - q_j\Bigr]^2,
\end{split}
\label{eq:Lphys}
\end{equation}
where subscript $j$ denotes evaluation at $(x_j,t_j;\theta)$. The implementation
evaluates the same residual in normalised coordinates and rescales the residual
terms for stable optimisation.

\textbf{Boundary condition loss} (enforce Eq.~\ref{eq:BC} at domain boundaries):
\begin{equation}
\begin{split}
\mathcal{L}_{\text{BC}}(\theta)
  &= \bigl[\hat{u}(0;\theta)\bigr]^2
   + \bigl[\hat{u}_x(0;\theta)\bigr]^2 \\
  &\quad + \bigl[\hat{u}_{xx}(L;\theta)\bigr]^2
   + \bigl[\hat{u}_{xxx}(L;\theta)\bigr]^2.
\end{split}
  \label{eq:Lbc}
\end{equation}
The zero-valued moment and shear conditions in Eq.~\eqref{eq:BC} are therefore
enforced through the corresponding derivative residuals.

All neural-network partial derivatives in Eqs.~\eqref{eq:Lphys}--\eqref{eq:Lbc}
are computed by PyTorch automatic differentiation \citep{Paszke2019}, avoiding
finite-difference derivative approximations at the collocation points. The PDE residual
$\mathcal{L}_\text{PDE}$ evaluated at 200 collocation points provides dense
structural information that functions as a physics-based data augmentation.

\subsubsection{Exponential Moving Average Adaptive Loss Weighting}
\label{sec:theory_ema}

Fixed loss weights caused one component to dominate in 3/10 configurations.
This study uses EMA-adaptive weighting with running averages
\begin{equation}
\bar{\mathcal{L}}_i^{(t)} = (1-\alpha)\,\bar{\mathcal{L}}_i^{(t-1)}
  + \alpha\,\mathcal{L}_i^{(t)},
\label{eq:ema}
\end{equation}
and dynamic weights $\lambda_i^{(t)} = \bar{\mathcal{L}}_\text{avg}^{(t)}/\bigl(\bar{\mathcal{L}}_i^{(t)}+\varepsilon\bigr)$,
with $\alpha{=}0.02$, $\varepsilon{=}10^{-6}$, which balances component scales
by down-weighting persistently large-magnitude losses and up-weighting
smaller-magnitude components.

\subsection{First-Order Reliability Method}
\label{sec:theory_form}

Given a limit-state function $g(\mathbf{X}) \leq 0$ defining failure and $n$
standard-normal variables $U_i = (X_i - \mu_i)/\sigma_i$, with
$\mathbf{X}=T^{-1}(\mathbf{u})$ denoting the transformation from standard
normal space to physical variables, FORM defines the reliability index
\begin{equation}
\beta = \min_{\mathbf{u}:\,g(T^{-1}(\mathbf{u}))=0} \|\mathbf{u}\|_2
\label{eq:beta}
\end{equation}
and approximates failure probability as $P_f \approx \Phi(-\beta)$
\citep{Wang2022}. The Most Probable Point is located via SLSQP in
5--15 function evaluations. In the benchmarked implementation, SLSQP evaluates
user-defined algebraic limit-state functions and estimates the required local
sensitivities numerically by finite differences. Coupling the optimiser
directly to a trained PINN displacement evaluator is compatible with the same
interface but is not separately benchmarked in this paper.

\section{Methodology}
\label{sec:method}

\subsection{System Architecture Overview}
\label{sec:method_arch}

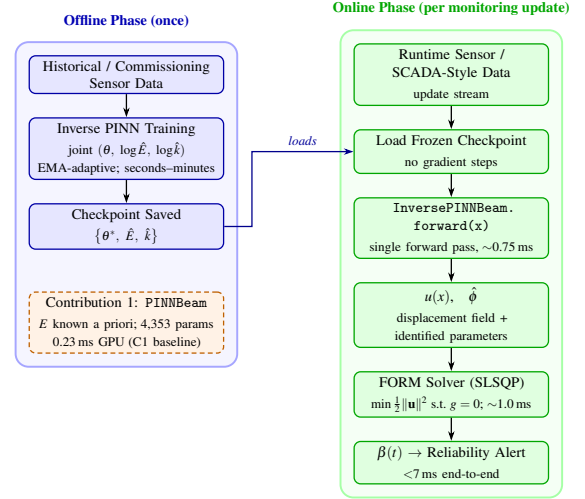
\begin{figure}[!t]
\centering
\resizebox{0.95\columnwidth}{!}{%
\begin{tikzpicture}[
  font=\small,
  node distance=0.5cm and 2.8cm,
  offbox/.style={draw=blue!65!black, fill=blue!10, rounded corners=4pt, thick,
                 minimum width=4.2cm, text width=4.0cm, align=center,
                 minimum height=0.72cm, inner sep=3pt},
  onbox/.style={draw=green!65!black, fill=green!12, rounded corners=4pt, thick,
                minimum width=4.2cm, text width=4.0cm, align=center,
                minimum height=0.72cm, inner sep=3pt},
  ctrbox/.style={draw=orange!75!black, fill=orange!10, rounded corners=4pt, thick,
                 densely dashed, minimum width=4.2cm, text width=4.0cm, align=center,
                 minimum height=0.72cm, inner sep=3pt},
  arr/.style={-{Stealth[length=6pt,width=4pt]}, thick},
  xarr/.style={-{Stealth[length=6pt,width=4pt]}, thick, blue!55!black},
  phasebox/.style={draw, rounded corners=8pt, very thick, inner sep=8pt},
]
\node[offbox] (hist)  {Historical / Commissioning\\Sensor Data};

\node[offbox, below=of hist] (inv_train)
  {Inverse PINN Training\\[2pt]
   {\footnotesize joint $(\boldsymbol{\theta},\;\log\hat{E},\;\log\hat{k})$}\\
   {\footnotesize EMA-adaptive; seconds--minutes}};

\node[offbox, below=of inv_train] (ckpt)
  {Checkpoint Saved\\[2pt]
   {\footnotesize $\bigl\{\boldsymbol{\theta}^{*},\;\hat{E},\;\hat{k}\bigr\}$}};

\draw[arr] (hist) -- (inv_train);
\draw[arr] (inv_train) -- (ckpt);

\node[ctrbox, below=0.9cm of ckpt] (fwd_bench)
  {Contribution~1: \texttt{PINNBeam}\\[2pt]
   {\footnotesize $E$ known a priori;\;4{,}353 params}\\
   {\footnotesize 0.23\,ms GPU\;(C1 baseline)}};

\node[onbox, right=of hist] (rt_data)
  {Runtime Sensor / SCADA-Style Data\\[2pt]
   {\footnotesize update stream}};

\node[onbox, below=of rt_data] (frozen)
  {Load Frozen Checkpoint\\[2pt]
   {\footnotesize no gradient steps}};

\node[onbox, below=of frozen] (fwd_pass)
  {\texttt{InversePINNBeam.}\\[0pt]
   \texttt{forward(x)}\\[2pt]
   {\footnotesize single forward pass,\;${\sim}$0.75\,ms}};

\node[onbox, below=of fwd_pass] (ux_ehat)
  {$u(x)$,\quad$\hat{\phi}$\\[2pt]
   {\footnotesize displacement field + identified parameters}};

\node[onbox, below=of ux_ehat] (form)
  {FORM Solver (SLSQP)\\[2pt]
   {\footnotesize $\min\tfrac{1}{2}\|\mathbf{u}\|^2$\;s.t.\;$g=0$;\;${\sim}$1.0\,ms}};

\node[onbox, below=of form] (beta_rul)
  {$\beta(t)$ $\to$ Reliability Alert\\[2pt]
   {\footnotesize $<$7\,ms end-to-end}};

\draw[arr] (rt_data)  -- (frozen);
\draw[arr] (frozen)   -- (fwd_pass);
\draw[arr] (fwd_pass) -- (ux_ehat);
\draw[arr] (ux_ehat)  -- (form);
\draw[arr] (form)     -- (beta_rul);

\draw[xarr] (ckpt.east) -- ++(0.6,0)
    |- node[near end, above, font=\footnotesize\itshape,
            text=blue!60!black]{loads}
    (frozen.west);

\begin{scope}[on background layer]
  \node[phasebox, fill=blue!4, draw=blue!35,
        fit=(hist)(inv_train)(ckpt)(fwd_bench)] (offbg) {};
  \node[phasebox, fill=green!4, draw=green!35,
        fit=(rt_data)(frozen)(fwd_pass)(ux_ehat)(form)(beta_rul)] (onbg) {};
\end{scope}

\node[above=5pt of offbg, font=\small\bfseries, text=blue!70!black]
  {Offline Phase (once)};
\node[above=5pt of onbg, font=\small\bfseries, text=green!65!black]
  {Online Phase (per monitoring update)};

\end{tikzpicture}%
}
\caption{Synthetic reliability-aware PINN workflow architecture.
  \textbf{Left (Offline Phase, once):} the Inverse PINN is trained on historical
  or commissioning sensor data to jointly identify the displacement surrogate
  $\boldsymbol{\theta}^*$ and physical parameters $\hat{E}$, $\hat{k}$ via
  EMA-adaptive co-optimisation (seconds to a few minutes in the present
  experiments); the checkpoint is saved.
  The Forward PINN (\texttt{PINNBeam}, Contribution~1) is validated independently
  as a 0.23\,ms C1-baseline benchmark when $E$ is known a priori; it is
  \emph{not} part of the online inference architecture.
  \textbf{Right (Online Phase, per monitoring update):} the frozen checkpoint is loaded
  and queried via a single forward pass of \texttt{InversePINNBeam} ($\sim$0.75\,ms),
  yielding $u(x)$ and identified parameters simultaneously; FORM then computes the
  reliability index $\beta(t)$ ($\sim$1.0\,ms for the OWT cantilever case);
  no gradient steps occur.
  Total end-to-end latency: $<$7\,ms, corresponding to $>$143\,Hz theoretical
  processing headroom.}
\label{fig:architecture}
\end{figure}

Fig.~\ref{fig:architecture} illustrates the two-phase DigiTurbine monitoring workflow.

The system operates in two phases. \textbf{Offline}: the Inverse PINN is trained
on historical or commissioning sensor data (seconds to a few minutes in the
present experiments); all model weights $\theta^*$
and identified parameters $\hat{\phi}$ are saved to a checkpoint.
\textbf{Online}: at every monitoring update the frozen checkpoint is loaded and a
single forward pass ($\sim$0.75\,ms) yields both the full displacement field $u(x)$
and the stored parameter estimates. FORM then evaluates the selected structural limit
state using these updated state estimates; the reported OWT cantilever case
requires about 1.0\,ms. No retraining occurs during online operation; periodic offline
retraining (e.g.\ after a storm event or on a monthly schedule) refreshes
$\hat{\phi}$ using new sensor history.  When $E$ is known in advance,
the lightweight Forward PINN (\texttt{PINNBeam}, 4{,}353 parameters) provides
full-field reconstruction at 0.23\,ms for the C1 baseline and serves as the
standalone Contribution~1 benchmark.

The software implementation uses a modular backend comprising
\texttt{PINNBeam}, \texttt{InversePINNBeam}, \texttt{FORMSolver},
and a Streamlit real-time monitoring dashboard.

\subsection{Forward PINN Design and Training}
\label{sec:method_fwd}

\subsubsection{Network Architecture}
\label{sec:method_net}

The network is a 2-hidden-layer MLP with 64 neurons per layer:
\[
\hat{u} = \mathcal{N}_\theta:\;
  \mathbb{R} \ni x \;\longmapsto\; \hat{u} \in \mathbb{R}.
\]
The $\tanh$ activation is used because it provides $\mathcal{C}^\infty$ smoothness,
and in particular $\mathcal{C}^4$ continuity in the $x$ direction needed for
accurate computation of $\partial^4\hat{u}/\partial x^4$ via automatic
differentiation. ReLU-type activations are unsuitable for this fourth-order
PINN because their higher derivatives are zero or ill-defined over much of the
domain. Hidden layers use PyTorch default initialisation
(Kaiming uniform); the output layer uses small-norm initialisation
($\text{std}=10^{-3}$, zero bias) for stable initial predictions near zero.
The parameter count is $1{\times}64 + 64{\times}64 + 64{\times}1 = 4{,}224$
weights plus $64+64+1=129$ biases, for $4{,}353$ trainable parameters
(approximately 34\,KB serialised).

\subsubsection{Training Protocol}
\label{sec:method_train}

Training uses the Adam optimiser \citep{Kingma2015} with learning rate
$\eta = 1{\times}10^{-3}$ for $N = 500$ epochs, full-batch over all
collocation points. Loss weights are computed adaptively from epoch~1 by the
EMA scheme (Section~\ref{sec:theory_ema}); no fixed initial weights are set.
Preliminary experiments with fixed weights
($\lambda_\text{PDE} = \lambda_\text{BC} = 100$,
$\lambda_\text{data} = 10$) led to poor convergence in 3 of 10 configurations,
motivating the fully adaptive approach.

Collocation setup: 200 collocation points uniformly distributed over $x \in [0,L]$
for $\mathcal{L}_\text{PDE}$; boundary conditions enforced pointwise at $x=0$ and $x=L$
for $\mathcal{L}_\text{BC}$; $N_d \in [50,100]$ labelled measurement points for
$\mathcal{L}_\text{data}$. All inputs are normalised to $[0,1]$; outputs are
normalised by the analytical peak displacement. The primary validation experiments
(C1--C10, I1--I4) use noiseless synthetic data; noise robustness is assessed
separately in I5--I8 at $\sigma_n \in \{1, 3, 5, 10\}\,\%$.

\subsection{Inverse PINN: Gradient Direction Problem and EMA-Adaptive Joint Co-Evolution}
\label{sec:method_inverse}

\subsubsection{The Gradient Direction Problem}
\label{sec:method_gd_problem}

In the standard simultaneous inverse PINN, network weights $\theta$ and unknown
physical parameters $\phi \in \{E, k_\text{soil}\}$ are treated as a single
vector of learnable variables, co-optimised via:

\begin{equation}
\begin{split}
(\hat{\theta},\hat{\phi})
  &= \arg\min_{\theta,\,\phi}\;
    \mathcal{L}_\text{data}(\theta) \\
  &\quad + \lambda_f\,\mathcal{L}_\text{PDE}(\theta,\phi)
         + \lambda_b\,\mathcal{L}_\text{BC}(\theta) \\
  &\quad + \lambda_\text{prior}\,\mathcal{L}_\text{prior}(\phi).
\end{split}
\label{eq:joint}
\end{equation}
Here $\mathcal{L}_\text{prior}$ is the Gaussian log-parameter penalty used in the
Bayesian-prior-informed runs; setting $\lambda_\text{prior}=0$ recovers the
no-prior ablations.

The gradient of the static PDE loss ($q - EI\hat{u}_{xxxx} = 0$) with respect
to $E$ is:

\begin{equation}
\frac{\partial \mathcal{L}_{\text{PDE}}}{\partial E}
= \frac{2}{N_f}\sum_j
  \bigl(E I \hat{u}_{xxxx,j} - q_j\bigr)
  \cdot I\hat{u}_{xxxx,j},
\label{eq:gradE}
\end{equation}

where $\hat{u}_{xxxx,j} = \partial^4\hat{u}/\partial x^4\big|_{x_j;\theta}$
\textit{depends entirely on the current network weights~$\theta$}. When $\theta$
is randomly initialised, $\hat{u}_{xxxx}$ is far from the physically correct
value, causing the gradient in Eq.~\eqref{eq:gradE} to provide systematically
misleading update directions for $E$. Simultaneously, $\mathcal{L}_\text{data}$
drives $\theta$ to fit observed displacements, but displacement-only data do
not uniquely constrain the fourth derivative needed for stiffness
identification. This circular dependency is:

\begin{align*}
\nabla_E \mathcal{L}_\text{PDE}
  &\;\text{requires physically consistent }\hat{u}^{(4)}_\theta, \\
\nabla_\theta \mathcal{L}_\text{PDE}
  &\;\text{shapes }\hat{u}^{(4)}_\theta\text{ around the current }E.
\end{align*}

Gradient-direction analysis (signed cosine similarity between
$\nabla_\Omega \mathcal{L}_\text{data}$ and $\nabla_\Omega \mathcal{L}_\text{PDE}$,
where $\Omega = \{\theta, \phi\}$ is the full parameter vector) confirmed
persistent opposition: mean cosine similarity $-0.21 \pm 0.07$ across 11 runs
(multiple seeds and noise levels), with ${\sim}80\,\%$ of training steps
producing opposing gradient directions. Notably,
$\partial\mathcal{L}_\text{data}/\partial E = 0$ because $E$ does not appear
in the forward prediction $\hat{u} = f_\theta(x)$; the conflict manifests
entirely through the shared network weights~$\theta$. A formal analysis is
provided in Appendix~\ref{app:proof}.

\subsubsection{EMA-Adaptive Joint Co-Evolution Algorithm}
\label{sec:method_2stage}

Algorithm~\ref{alg:twostage} presents the EMA-adaptive joint co-evolution
procedure, designed to mitigate the circular dependency by training network
weights~$\theta$ and physical parameters~$\phi$ \emph{jointly from epoch~1},
removing the need for any freeze or unfreeze step.

\begin{algorithm*}[htbp]
\caption{EMA-Adaptive Joint Co-Evolution for Inverse PINN Parameter Identification.}
\label{alg:twostage}
\begin{algorithmic}[1]
\Require Data $\mathcal{D} = \{(x_i, u_i^{\text{obs}})\}$; initial guess $\phi_0$;
         optional log-prior moments $(\mu_\phi,\sigma_\phi)$ and
         $\lambda_\text{prior}$; $\alpha{=}0.02$; $\gamma{=}10$; $N{=}5{,}000$.
\Ensure Identified $\hat{\phi}$.
\State $\theta \leftarrow$ Kaiming; $\phi \leftarrow \log(\phi_0)$; $\bar{\mathcal{L}}_i \leftarrow 0$
\For{$e = 1$ \textbf{to} $N$}
  \State Compute $\mathcal{L}_\text{data}(\theta)$, $\mathcal{L}_\text{PDE}(\theta,\phi)$, $\mathcal{L}_\text{BC}(\theta)$, and $\mathcal{L}_\text{prior}(\phi)$
  \State $\bar{\mathcal{L}}_i \leftarrow (1-\alpha)\bar{\mathcal{L}}_i + \alpha\mathcal{L}_i$;\enspace
         $\lambda_i \leftarrow \bar{\mathcal{L}}_\text{avg}/(\bar{\mathcal{L}}_i + \varepsilon)$
  \State $\mathcal{L} \leftarrow {\textstyle\sum_{i\in\{\text{data},\text{PDE},\text{BC}\}}} \lambda_i\mathcal{L}_i
         + \lambda_\text{prior}\mathcal{L}_\text{prior}$
  \State $\theta \leftarrow \theta - \eta_\theta\,\nabla_\theta\mathcal{L}$;\enspace
         $\phi \leftarrow \phi - \gamma\eta_\theta\,\nabla_\phi\mathcal{L}$
\EndFor
\State \Return $\hat{\phi} \leftarrow \exp(\phi)$
\end{algorithmic}
\end{algorithm*}

\textbf{Joint training rationale.} By training $\theta$ and $\phi$ simultaneously,
the network and unknown parameters remain \emph{mutually consistent} throughout
optimisation. EMA-based adaptive loss weighting is designed to keep component
magnitudes comparable without fixed manual tuning. In early epochs, the large
data residual drives the network toward the observed displacement field while
the balancing prevents that term from swamping the PDE, boundary, and prior
terms. In later epochs, as the data loss falls, the PDE and prior terms provide
the main parameter-identification signal for $\phi$.

\textbf{Fundamental limitation for 4th-order PDEs.}
Experimental investigation (Section~\ref{sec:results_inv}) reveals that
\emph{without} Bayesian priors, the gradient direction problem prevents
convergence for the Euler--Bernoulli beam equation with displacement-only data.
However, the addition of a weak Bayesian prior, centred at
the correct design/as-built value of $E$ in the reported synthetic cases and
given a wide uncertainty range
($\sigma = 1.0$ in log-space, covering approximately $\times 2.7$ in either
direction), substantially mitigates the circular dependency by anchoring the parameter search
near the physically plausible regime. This allows the EMA-adaptive mechanism to
drive convergence when the prior is accurately centred, achieving
$<$0.04\,\% identification error across all synthetic test configurations in
Section~\ref{sec:results_inv}.

\textbf{Why classical freeze-thaw also fails.} Freeze-thaw staging
(fix $\phi=\phi_0$, pretrain $\theta$, then optimise $\phi$)
converges only when Stage~0 has oracle access to $E^*$; with an inaccurate
non-oracle initial guess $\phi_0$, the network acquires derivatives consistent with
$\phi_0$ and Stage~1 remains trapped.

\textbf{Parameter learning-rate multiplier.} Setting $\eta_\phi = \gamma\eta_\theta$
with $\gamma=10$ makes physical parameters update faster than network weights;
this was important for convergence within the 5{,}000-epoch budget.
\subsection{FORM Reliability Layer}
\label{sec:method_form}

The implemented reliability layer uses FORM to evaluate benchmark structural
limit-state functions using the current structural state supplied by the
monitoring workflow. A generic displacement-style limit state is:

\begin{equation}
g(\mathbf{X}) = u_{\text{allow}} - u_{\text{model}}(x_\text{cr}, t;\,\mathbf{X}),
\label{eq:lsf}
\end{equation}

where $\mathbf{X}$ are normally distributed resistance, load, and geometry
variables with means from design specifications and CoV of 5--15\,\%, and
$u_\text{allow}$ is the code-prescribed allowable displacement \citep{DNV2021}.
For the reported validation and timing runs, $u_{\text{model}}$ or the equivalent
load effect is evaluated by analytical/algebraic benchmark functions; direct
PINN-in-the-loop FORM is an implementation extension rather than a benchmarked
result in this paper. The limit-state gradient $\nabla_{\mathbf{X}} g$ is
computed numerically by SLSQP's built-in finite-difference scheme.

The SLSQP optimisation (Eq.~\ref{eq:beta}) converges in 5--15 function evaluations for the present
problem class, yielding $\beta(t)$ in $\sim$1.0\,ms for the OWT cantilever
root-moment capacity case in the reported CPU timing run.

The end-to-end system operates in two phases.
\textbf{Offline training} (once, on commissioning or historical data, takes seconds to a few minutes):
the Inverse PINN is trained to simultaneously learn the displacement surrogate
$\theta^*$ and identify the physical parameters
$\hat{\phi} = \{\hat{E},\hat{k}\}$; the resulting checkpoint
$\{\theta^*, \log\hat{E},\log\hat{k}\}$ is saved.
\textbf{Online inference} (every monitoring update, frozen model):
a single forward pass through the frozen Inverse PINN yields the full
displacement field $u(x)$ and the stored parameter estimates simultaneously.
FORM then evaluates the selected algebraic/structural limit state using the
updated PINN-derived state estimate. The complete online pipeline is:
\[
\resizebox{\columnwidth}{!}{$
\underbrace{%
\text{Sensor data}
\xrightarrow{\;\text{Frozen Inverse PINN},\ {\sim}0.75\text{ms}\;}
\bigl[u(x),\;\hat{\phi}\bigr]
\xrightarrow{\;\text{FORM limit state},\ {\sim}1.0\text{ms}\;}
\beta(t)
}_{\text{online inference only; no gradient steps; }{<}7\text{\,ms total}}.
$}
\]
Note: The Forward PINN (Contribution~1) provides the 0.23\,ms C1-baseline
inference benchmark and is used when $E$ is known in advance; in the online inference loop the Inverse PINN subsumes this role, since
\texttt{InversePINNBeam} inherits from \texttt{PINNBeam} and produces $u(x)$
in the same forward pass that exposes $\hat{\phi}$.

\subsection{Validation Data Strategy and Hardware}
\label{sec:method_data}

All results use simulation-based synthetic data against known ground truth.
Sources include: (1)~\textit{analytical solutions} (Eq.~\ref{eq:analytical});
(2)~\textit{finite-difference Winkler-beam references}; and (3)~parametric
sweeps over $L = 10$--$15$\,m, $E = 100$--$250$\,GPa,
$q = 1{,}000$--$5{,}000$\,N/m. Preprocessing uses min-max normalisation;
Gaussian noise of 1--10\,\% is injected only in the specified noise-robustness
experiments, and models are evaluated against dense reference grids.
Hardware: Intel i7-12700H, NVIDIA RTX~A2000 (8\,GB), PyTorch 2.5.1.
Timing: 20 warm-up + 500 timed passes per device with
\texttt{torch.cuda.synchronize()}. All-configuration mean GPU inference is
0.381\,ms (range 0.230--0.500\,ms); mean CPU inference is 0.605\,ms
(0.391--0.767\,ms); both devices satisfy the 10\,ms real-time ceiling with
$>$16$\times$ margin. OpenFAST execution, field SCADA ingestion, and
aero-elastic Morison-driven validation are deferred to future field-calibration.

\section{Results}
\label{sec:results}

\subsection{Forward PINN Validation}
\label{sec:results_fwd}

\subsubsection{Experimental Configurations and Performance}
\label{sec:results_fwd_configs}

Ten synthetic benchmark configurations spanning network architecture, observation
density, beam length, load magnitude, and material stiffness
are detailed in Table~\ref{tab:forward_configs}.

\begin{table*}[htbp]
\caption{Forward PINN validation: 10 test configurations spanning network architecture,
collocation density, data quantity, beam geometry, load magnitude, and material stiffness.
RMSE and R$^2$ quantify prediction accuracy against analytical solutions.
GPU and CPU inference columns give means of 500 back-to-back passes timed in a single process
(RTX~A2000 with \texttt{cuda.synchronize()}; i7-12700H with \texttt{perf\_counter()}; 20 warm-up passes discarded).}
\label{tab:forward_configs}
\centering
\small
\resizebox{\textwidth}{!}{%
\begin{tabular}{llccccccccc}
\toprule
\textbf{ID} & \textbf{Configuration} & \textbf{Arch.} &
\textbf{$N_f$} & \textbf{$N_d$} & \textbf{RMSE (mm)} & \textbf{R$^2$} &
\textbf{Train (s)} & \textbf{GPU Inf.\ (ms)} & \textbf{CPU Inf.\ (ms)} & \textbf{Ratio} \\
\midrule
C1  & Baseline (standard)            & [64,64]    & 200 &  50 & 0.068 & 0.908 & 11.3 & 0.230 & 0.391 & 1.70 \\
C2  & Smaller network                & [32,32]    & 200 &  50 & 0.063 & 0.921 & 11.0 & 0.255 & 0.410 & 1.61 \\
C3  & Deeper network (4 layers)      & [64]$\times$4 & 200 &  50 & 0.058 & 0.933 & 22.5 & 0.355 & 0.683 & 1.93 \\
C4  & Wider network                  & [128,128]  & 200 &  50 & 0.061 & 0.925 & 11.4 & 0.252 & 0.438 & 1.74 \\
C5  & Dense collocation              & [64,64]    & 500 &  50 & 0.064 & 0.917 &  8.7 & 0.254 & 0.587 & 2.31 \\
C6  & More training data             & [64,64]    & 200 & 100 & 0.065 & 0.915 &  9.1 & 0.500 & 0.692 & 1.38 \\
C7  & Longer beam ($L$=15\,m)        & [64,64]    & 200 &  50 & 0.349 & 0.905 &  9.2 & 0.493 & 0.767 & 1.55 \\
C8  & Higher load ($q$=5{,}000\,N/m) & [64,64]    & 200 &  50 & 0.302 & 0.927 &  9.2 & 0.496 & 0.721 & 1.45 \\
C9  & Lower stiffness ($E$=100\,GPa) & [64,64]    & 200 &  50 & 0.138 & 0.913 &  9.0 & 0.499 & 0.608 & 1.22 \\
C10 & Challenging (long$+$high load) & [128,128]  & 300 &  75 & 0.183 & 0.999 &  9.4 & 0.470 & 0.750 & 1.60 \\
\midrule
\textbf{Mean} & & & & &
$\mathbf{0.135 \pm 0.109}$ & $\mathbf{0.926}$ & $\mathbf{11.1}$ & $\mathbf{0.381}$ & $\mathbf{0.605}$ & $\mathbf{1.65}$ \\
\textbf{Pass (R$^2>0.90$)} & & & & & & & & & & \textbf{10/10} \\
\bottomrule
\end{tabular}%
}
\end{table*}

\textbf{Accuracy.} All ten configurations achieve R$^2 > 0.90$, with mean RMSE
$0.135 \pm 0.109$\,mm. Six of ten configurations achieve sub-100\,$\mu$m RMSE.
Configuration C7 (longer beam, $L=15$\,m) shows the largest error (0.349\,mm)
because longer spans produce larger absolute displacements. The `challenging case'
C10 (long beam, high load, wider network) achieves the best R$^2 = 0.999$,
showing strong accuracy on the most demanding synthetic configuration tested.

\textbf{Inference speed.} Mean GPU inference is 0.381\,ms and mean CPU
inference is 0.605\,ms over the ten configurations (500-rep joint benchmark;
GPU $1.65\times$ faster than CPU; fastest config 0.230\,ms on GPU). GPU is
26$\times$, and CPU is 17$\times$, below the 10\,ms real-time target. Compared
with high-fidelity offline FEA or OpenFAST-style analysis, this is an online
PINN/state-estimation query rather than an equivalent full-model solve; the
speed comparison is therefore indicative rather than a measured FEA replacement
ratio.

\begin{figure}[htbp]
\centering
\includegraphics[width=0.95\linewidth]{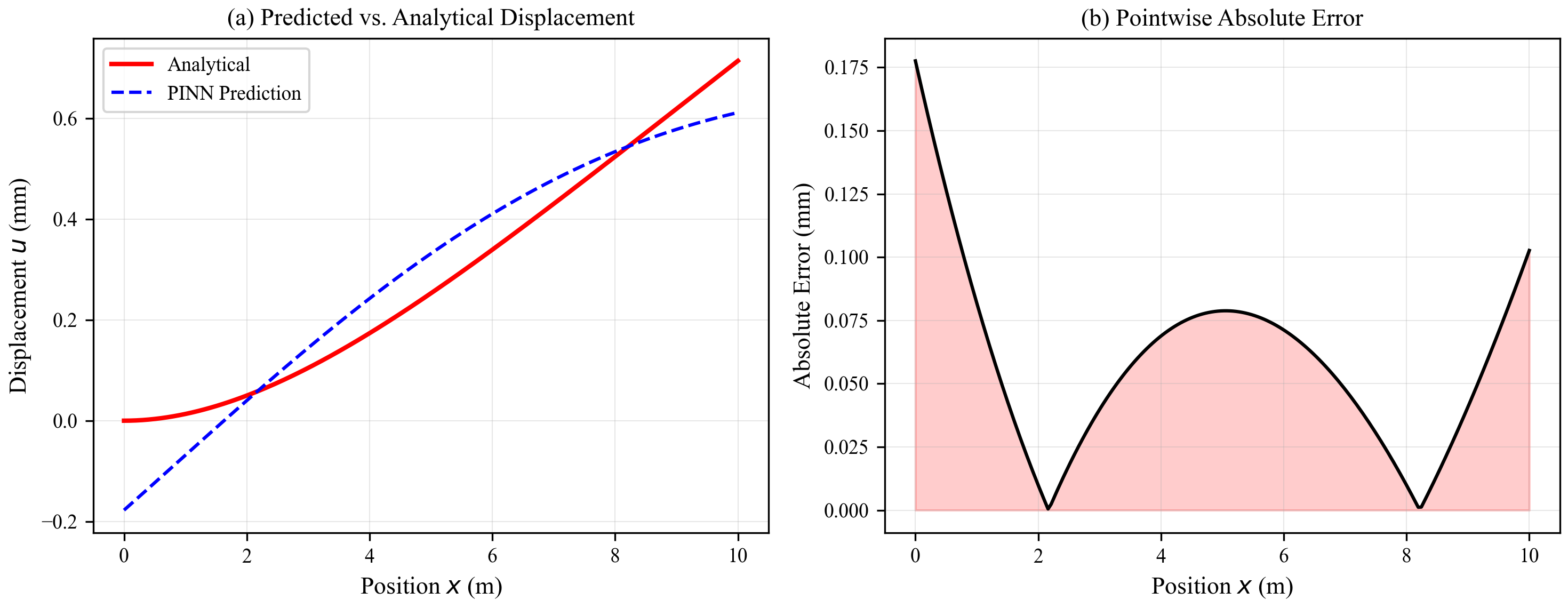}
\caption{Representative PINN displacement field prediction (C1 baseline, 0\,\%
noise). Left: PINN prediction vs.\ analytical solution across the full beam span.
Right: pointwise absolute error; maximum deviation 0.178\,mm at the clamped end
($x = 0$), reflecting the soft boundary-condition enforcement used in the
baseline PINN.}
\label{fig:fwd_pred}
\end{figure}

\textbf{Data efficiency.} Accurate predictions from $N_d = 50$ training points
(C1--C5, C7--C9) support the physics-regularisation hypothesis: PDE residuals at
200 collocation points provide dense structural information that effectively
compensates for scarce measurements. The baseline configuration achieves
RMSE $= 0.068$\,mm and R$^2 = 0.908$ from just 50 data points.
C10, using 75 data points, achieves R$^2 = 0.999$.
Figure~\ref{fig:fwd_pred} shows the baseline prediction against the analytical reference.

\begin{figure*}[htbp]
\centering
\includegraphics[width=0.48\linewidth]{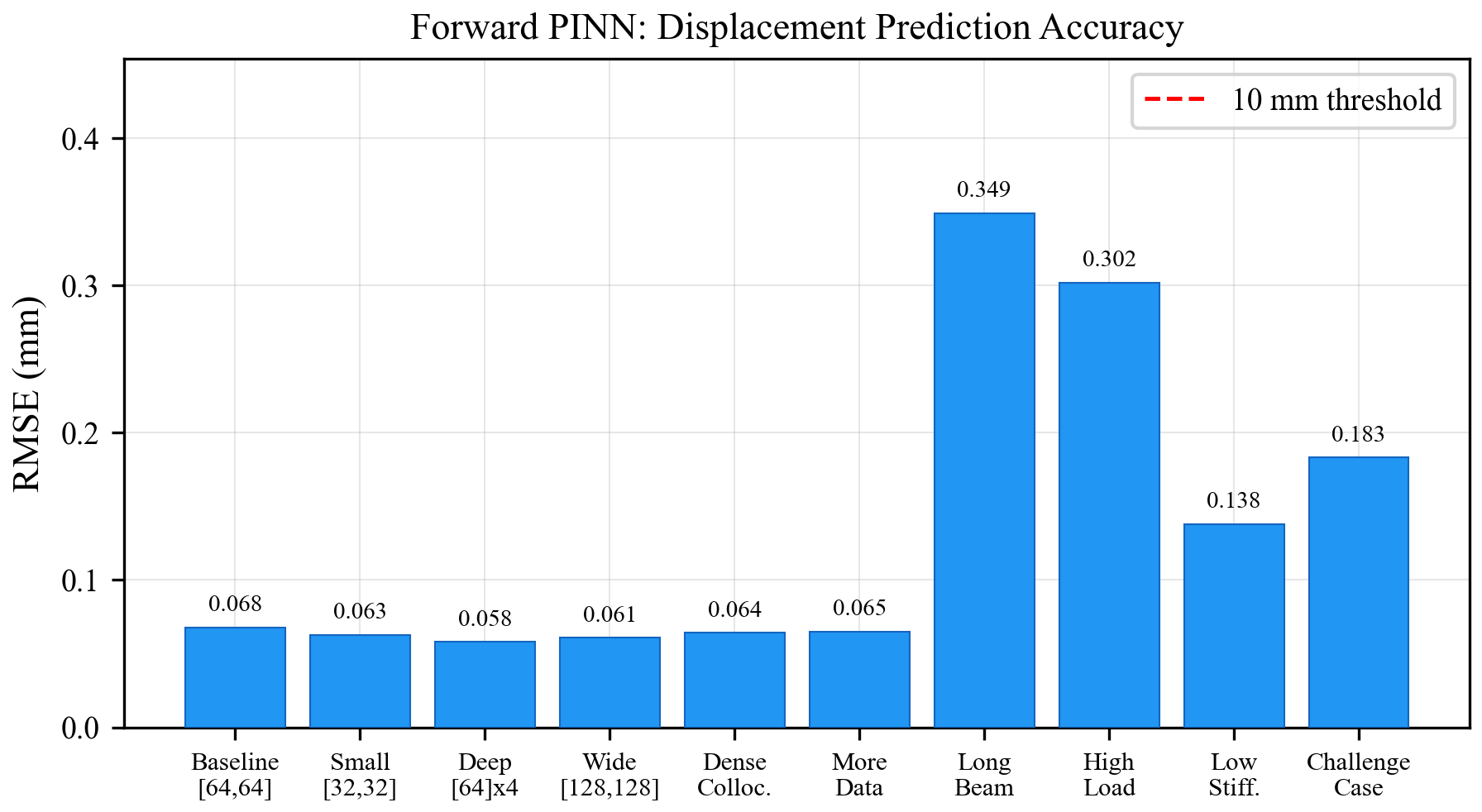}\hfill
\includegraphics[width=0.48\linewidth]{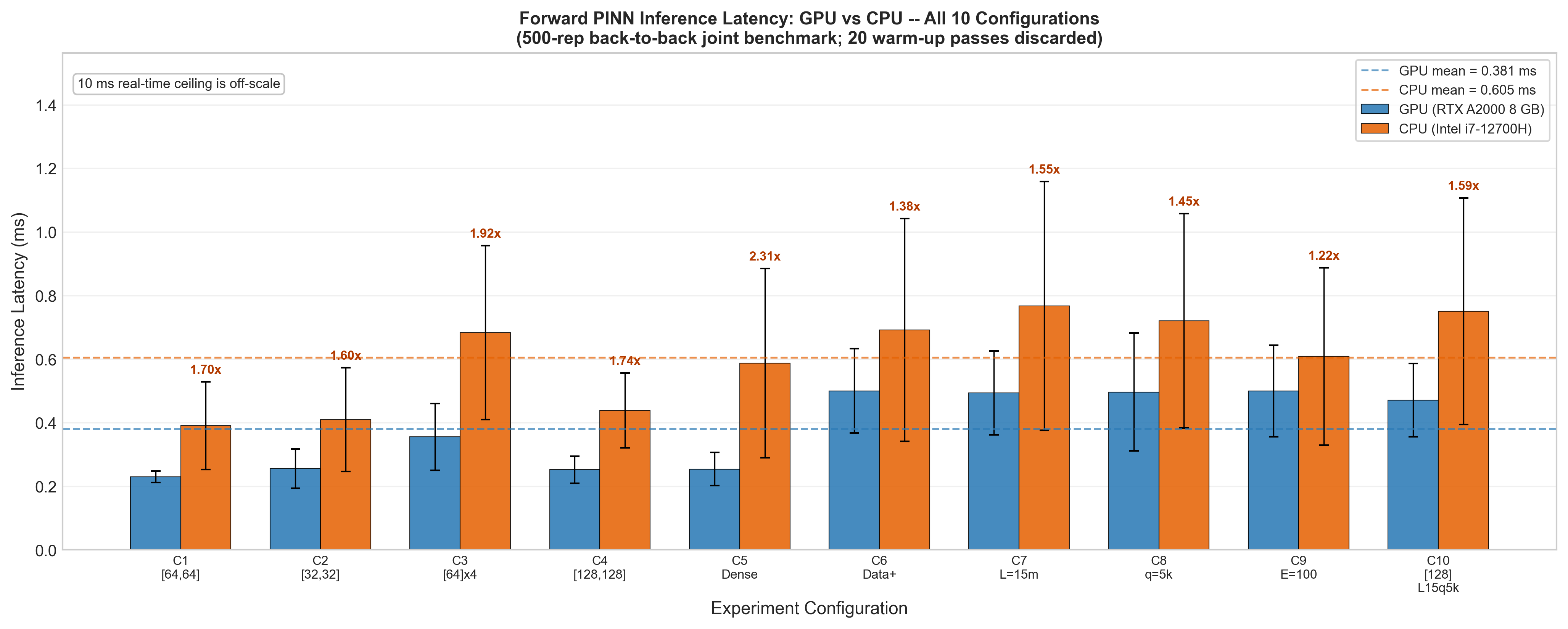}
\caption{\textbf{Forward PINN performance.} Left: RMSE for all ten configurations
(all $<$0.35\,mm; 10/10 pass R$^2>0.90$). Right: GPU vs.\ CPU inference latency
for all ten configurations (RTX~A2000 8\,GB vs.\ i7-12700H; 500-rep joint benchmark;
mean GPU 0.381\,ms, mean CPU 0.605\,ms; GPU $>$26$\times$ and CPU $>$16$\times$ faster than
10\,ms real-time target).}
\label{fig:fwd_perf}
\end{figure*}

\textbf{Architecture sensitivity.} Among the tested architectures, the
2-hidden-layer [64,64] baseline gives the best accuracy--speed trade-off:
accuracy comparable to [32,32], lower training cost than the 4-layer network,
and no clear gain from [128,128].
Figure~\ref{fig:fwd_perf} summarises RMSE and inference latency across the ten configurations.

\subsection{Inverse PINN: Gradient Direction Problem and EMA-Adaptive Joint Co-Evolution}
\label{sec:results_inv}

\subsubsection{Baseline Failure}
\label{sec:results_inv_baseline}

Table~\ref{tab:baseline} documents failure of standard simultaneous
inverse PINN training in the tested large-offset Young's modulus identification cases
($E_\text{true} = 210$\,GPa, $E_0 = 100$\,GPa in all cases).

\begin{table*}[htbp]
\caption{Standard (baseline) simultaneous inverse PINN: failure in the tested
large-offset cases.
$E_\text{true} = 210$\,GPa; $E_0 = 100$\,GPa. Parameters moved away from
truth in all four experiments.}
\label{tab:baseline}
\centering
\small
\begin{tabular}{lcccc}
\toprule
\textbf{Exp.} & \textbf{Noise} & $E_\text{final}$ (GPa) &
\textbf{Error (\%)} & \textbf{Pass ($<5\,\%$)} \\
\midrule
B1 & 0\,\%  & 70.1 & 66.6 & $\times$ \\
B2 & 0\,\%$^*$ & 78.1 & 62.8 & $\times$ \\
B3 & 0\,\%$^\dagger$  & 76.5 & 63.6 & $\times$ \\
B4 & 10\,\% & 78.8 & 62.5 & $\times$ \\
\midrule
\textbf{Mean} & & $75.9 \pm 4.0$ & $\mathbf{63.9 \pm 1.9}$ & \textbf{0/4 (0\,\%)} \\
\bottomrule
\multicolumn{5}{l}{$^*$B2: simultaneous $E{+}k$ identification.}\\
\multicolumn{5}{l}{$^\dagger$B3: multi-parameter ($E$, $k$, damage).}\\
\end{tabular}
\end{table*}

Table~\ref{tab:inverse_experiments} summarises the inverse benchmark categories
examined in this study, showing the main goals and the distinguishing setup for
each class of experiment.

\begin{table}[!t]
\caption{Inverse PINN benchmark categories used in this study.}
\label{tab:inverse_experiments}
\centering
\small
\setlength{\tabcolsep}{4pt}
\begin{tabularx}{\columnwidth}{@{}>{\raggedright\arraybackslash}p{0.24\columnwidth} >{\raggedright\arraybackslash}X >{\raggedright\arraybackslash}p{0.22\columnwidth}@{}}
\toprule
\textbf{Category} & \textbf{Goal} & \textbf{Key setup} \\
\midrule
Baseline failure & identify $E$ and/or $k$ with standard joint PINN & no prior, large initial offset \\
Multi-parameter test & simultaneous $E{+}k$ identification & coupled parameters, 0\% noise \\
Noise robustness & assess stability under noisy observations & 1--10\% measurement noise \\
Bayesian mitigation & recover parameters with weak prior information & log-normal prior on $\log E$ \\
\bottomrule
\end{tabularx}
\end{table}

\begin{figure}[htbp]
\centering
\includegraphics[width=0.95\linewidth]{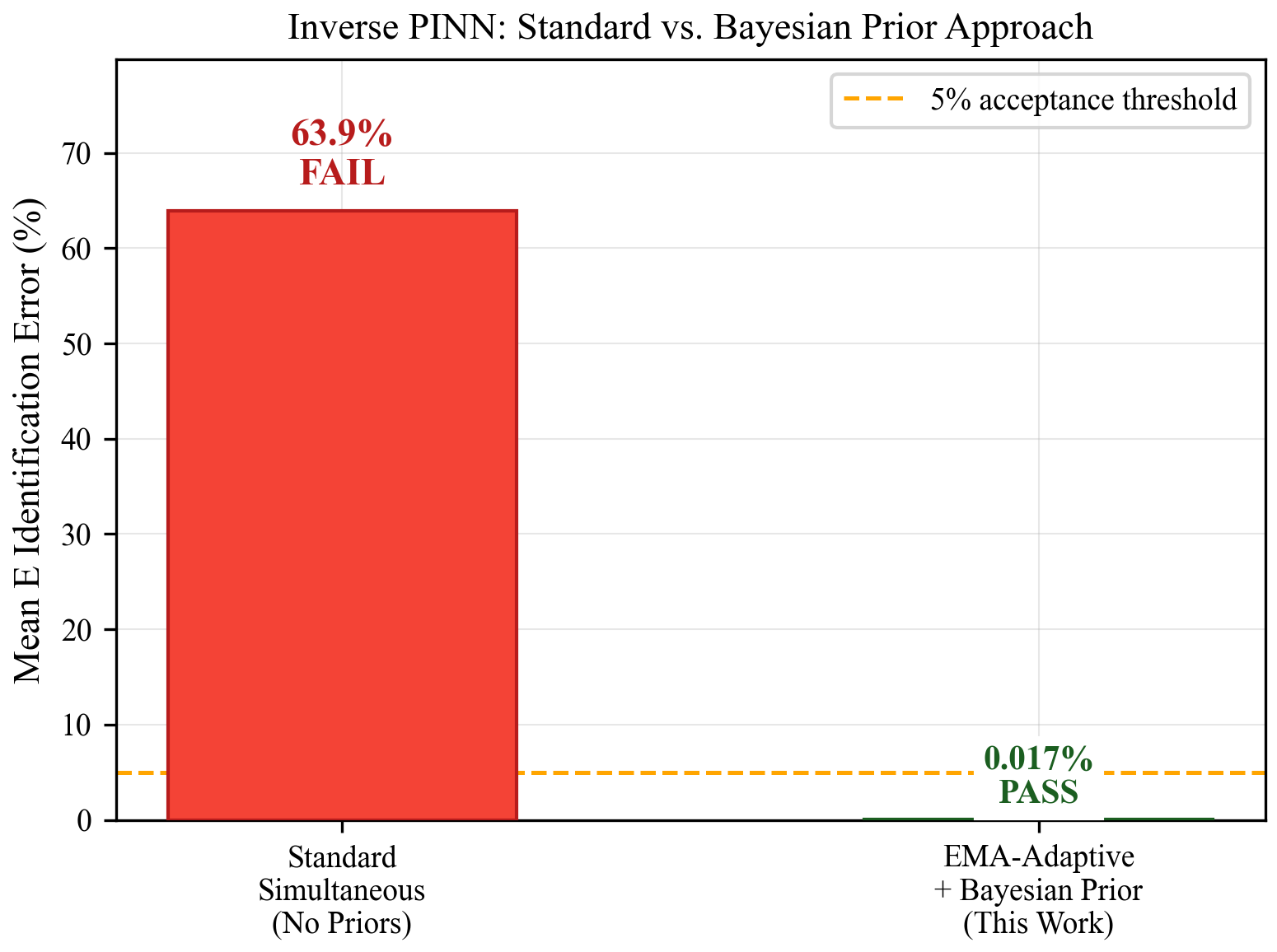}
\caption{Baseline inverse PINN: parameter identification error across the four
tested baseline experiments.  Each tested configuration exhibits large parameter error ($>$50\,\%),
with $E$ diverging \emph{away} from the true value.}
\label{fig:inv_baseline_failures}
\end{figure}

$E$ moved \emph{away} from the true value throughout all baseline training runs,
confirming the gradient direction problem. Gradient-direction analysis of the
full parameter vector $\Omega = \{\theta, \phi\}$ across all training steps
yielded mean cosine similarity $-0.21 \pm 0.07$ (11 runs, multiple seeds),
with ${\sim}80\,\%$ of steps producing opposing gradient directions, confirming
persistent gradient conflict. Figure~\ref{fig:inv_baseline_failures} illustrates
the baseline parameter divergence.

\subsubsection{EMA-Adaptive Joint Co-Evolution with Bayesian Priors: Mitigation}
\label{sec:results_inv_2stage}

\begin{figure}[htbp]
\centering
\includegraphics[width=0.95\linewidth]{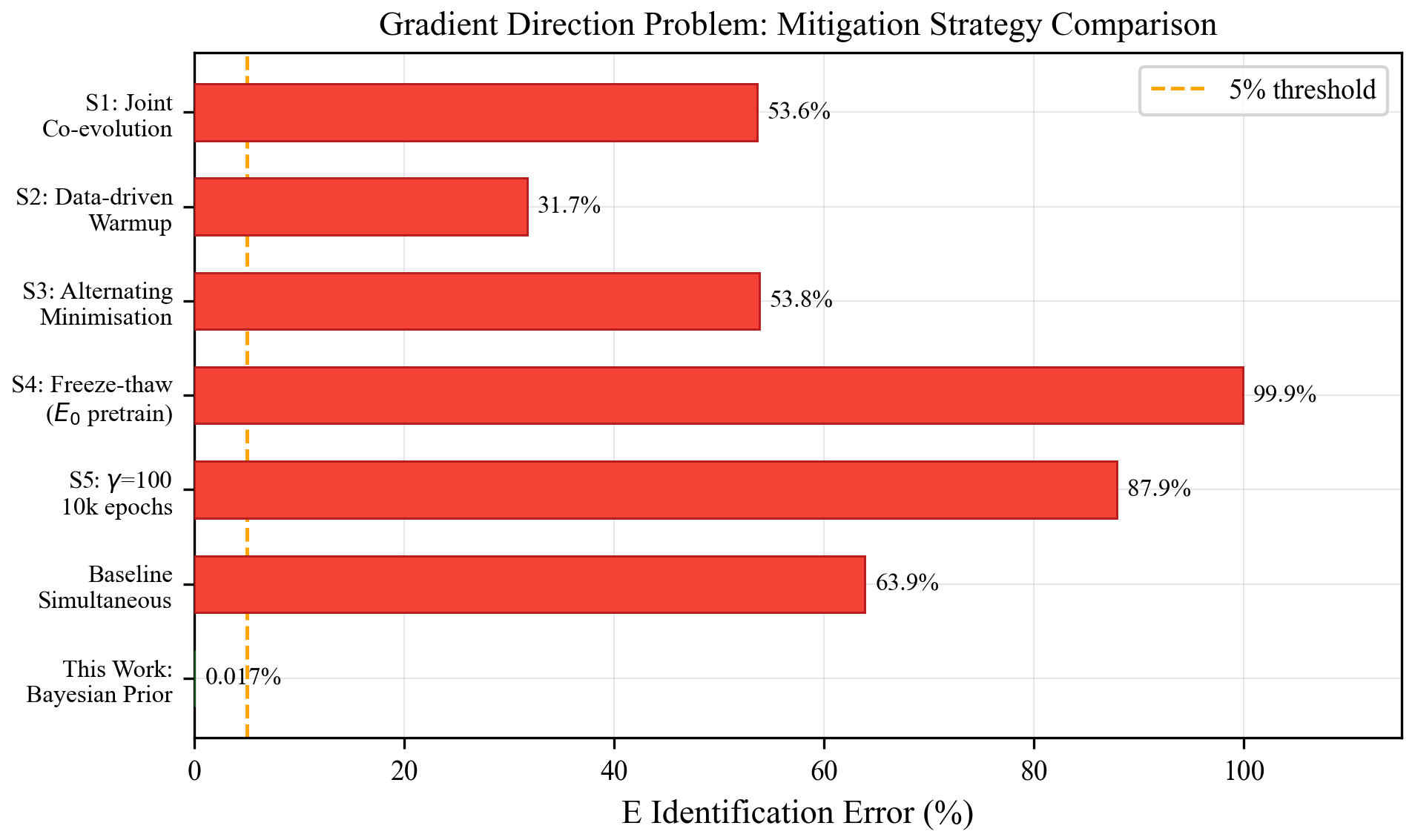}
\caption{Gradient direction problem: comparison of mitigation strategies.
Five approaches without Bayesian priors fail in the tested large-offset setup
($>$5\,\% error).
The EMA-adaptive joint co-evolution with a weak Bayesian prior
substantially mitigates the problem (mean $E$ error 0.017\,\%).}
\label{fig:inv_strategy_comparison}
\end{figure}

Table~\ref{tab:twostage_nopri} first presents the results of five mitigation strategies
applied \emph{without} Bayesian priors (pure loss-weighting and scheduling modifications),
all of which fail in this tested large-offset setup. Table~\ref{tab:twostage} then presents the full EMA-adaptive
joint co-evolution with a weak Bayesian prior, which substantially mitigates the gradient direction
problem and achieves 8/8 success across all eight tested configurations. Figure~\ref{fig:inv_strategy_comparison}
compares the mitigation strategies visually.

\begin{table*}[htbp]
\caption{Mitigation strategies \emph{without} Bayesian priors: all tested
variants fail.
$E_\text{true} = 210$\,GPa; $E_0 = 100$\,GPa; 5{,}000 epochs.
The gradient direction problem persists for 4th-order PDEs when only loss-weighting
or scheduling modifications are applied.}
\label{tab:twostage_nopri}
\centering
\small
\begin{tabular}{lcccc}
\toprule
\textbf{Strategy} & $\hat{E}$ (GPa) & \textbf{Error (\%)} &
\textbf{Direction} & \textbf{Pass ($<5\,\%$)} \\
\midrule
S1: Joint co-evolution (no prior) & 97.3 & 53.6 & Away & $\times$ \\
S2: Data-driven warmup + joint & 143.4 & 31.7 & Toward & $\times$ \\
S3: Alt.\ minimisation (10 outer) & 97.0 & 53.8 & Stuck & $\times$ \\
S4: Freeze-thaw ($E_0$ pretrain) & $<1.0$ & 99.9 & Collapse & $\times$ \\
S5: $\gamma=100$, 10k epochs & 25.4 & 87.9 & Diverge & $\times$ \\
\midrule
\textbf{Baseline (simultaneous)} & 75.9 & 63.9 & Away & $\times$ \\
\bottomrule
\end{tabular}
\end{table*}

\textbf{Why pure loss-weighting fails.}
Strategy~S2 (data-driven warmup: 20\,\% epochs data-only pretraining, then joint)
achieves partial improvement by learning the displacement shape from measurement
data before enabling parameter updates, correctly moving $E$ \emph{toward} the
true value ($100 \to 143$\,GPa). However, convergence stalls at 31.7\,\% error
because the network's 4th derivative $\hat{u}^{(4)}$ is still governed entirely by
the PDE constraint at the current (incorrect) $E$.
Strategy~S4 (freeze-thaw with $E_0$ instead of oracle $E^*$) collapses:
the PDE loss $\mathcal{L}_\text{PDE}
= (EI\hat{u}_{xxxx} - q)^2$ is minimised at $E = 0$ when $\hat{u}_{xxxx}$ has the
wrong magnitude.

\textbf{Mitigation: Bayesian-prior-informed joint co-evolution.}
Adding a weak Bayesian prior on $\log E$ (prior mean at the correct
design/as-built specification in these synthetic cases, $\sigma = 1.0$ in
log-space, corresponding to a $\times 2.7$ uncertainty range)
provides a regularisation signal that anchors the parameter near the physically
plausible regime. This breaks the circular dependency: instead of the PDE alone
determining $\hat{u}^{(4)}$, the prior prevents $E$ from drifting to spurious values,
allowing the EMA-adaptive mechanism to find the correct $E$.

Table~\ref{tab:twostage} presents the complete validation results.

\begin{table*}[htbp]
\caption{EMA-Adaptive Joint Co-Evolution with Bayesian prior: 8/8 tested
configurations pass.
$E_\text{true} = 210$\,GPa; $E_0 = 105$\,GPa (50\,\% initial error); 5{,}000 epochs.
Prior: $\log E \sim \mathcal{N}(\log E_\text{design},\; \sigma^2 = 1.0)$, a weak engineering
prior centered at the as-built design specification; prior weight $10^{-2}$.
Experiments I3 and I4 (simultaneous $E+k$) used adapted settings for
multi-parameter difficulty: $E_0 = 168$\,GPa (20\,\% error),
7{,}000 epochs, $\sigma_E = 0.3$, $N_d = 100$.}
\label{tab:twostage}
\centering
\small
\begin{tabular}{llccccc}
\toprule
\textbf{ID} & \textbf{Configuration} & \textbf{Noise} &
\textbf{E Error (\%)} & \textbf{$k$ Error (\%)} &
\textbf{RMSE (mm)} & \textbf{Pass} \\
\midrule
I1 & $E$ identification only           & 0\,\%   & 0.019 & --     & 0.333 & \checkmark \\
I2 & $k$ identification ($E$ known)    & 0\,\%   & 0.0    & 0.019 & 0.032 & \checkmark \\
I3 & $E + k$ simultaneous              & 0\,\%   & 0.008 & 0.001 & 0.045 & \checkmark \\
I4 & $E + k$ with noise                & 10\,\%  & 0.025 & 0.0002 & 0.064 & \checkmark \\
I5 & $E$ identification, 1\,\% noise & 1\,\%   & 0.025 & --     & 0.328 & \checkmark \\
I6 & $E$ identification, 3\,\% noise & 3\,\%   & 0.017 & --     & 0.338 & \checkmark \\
I7 & $E$ identification, 5\,\% noise & 5\,\%   & 0.005 & --     & 0.346 & \checkmark \\
I8 & $E$ identification, 10\,\% noise & 10\,\% & 0.037 & --     & 0.387 & \checkmark \\
\midrule
\textbf{Mean} & & & $\mathbf{0.017}$ & $\mathbf{0.007}$ &
$\mathbf{0.234}$ & \textbf{8/8} \\
\bottomrule
\end{tabular}
\end{table*}

\begin{figure}[htbp]
\centering
\includegraphics[width=0.95\linewidth]{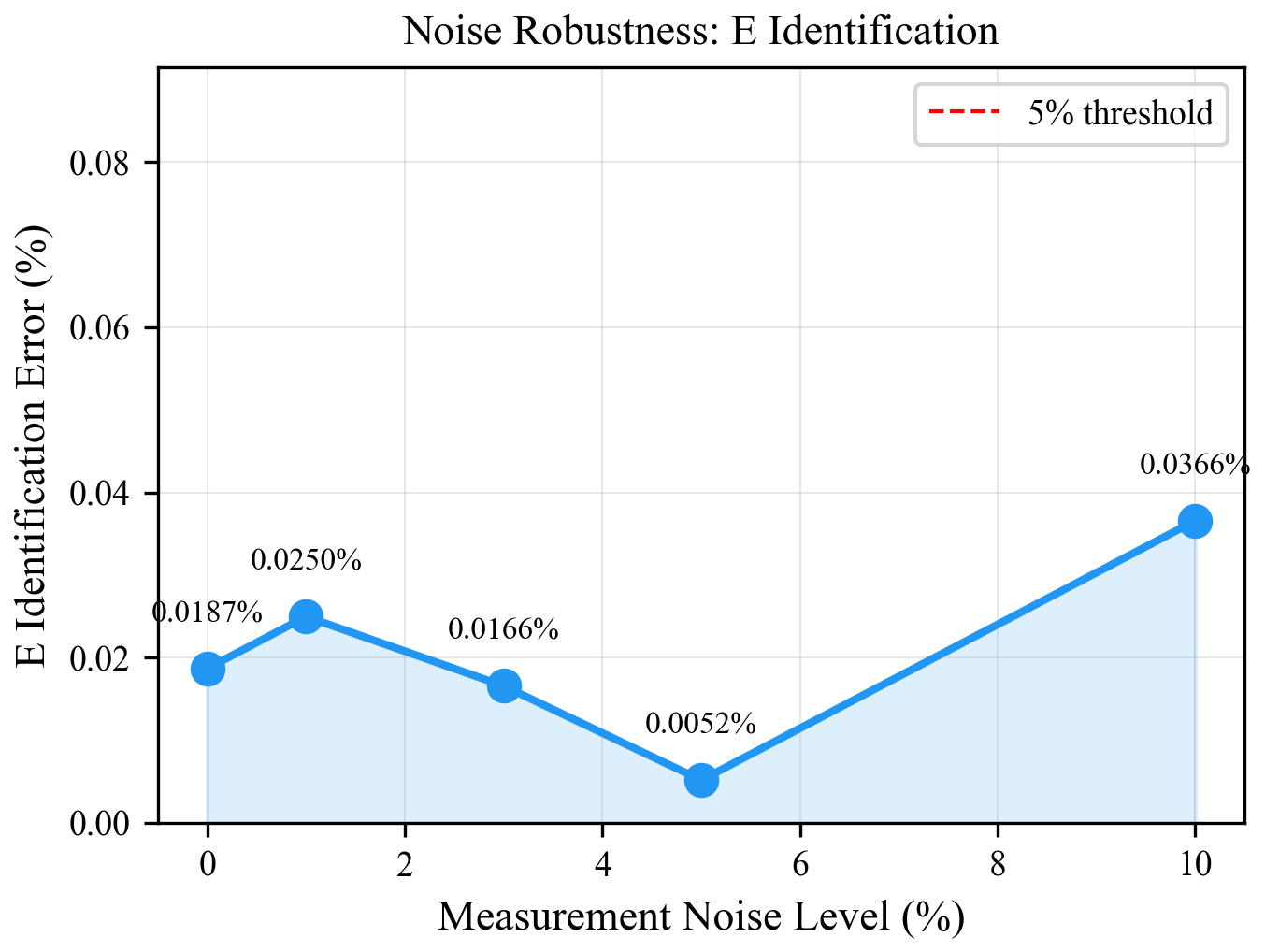}
\caption{Noise robustness of the Bayesian-prior-informed inverse PINN.
$E$ identification error remains below 0.04\,\% for the tested noise levels up
to 10\,\%.}
\label{fig:inv_noise_robust}
\end{figure}

\textbf{Key findings.}
(1)~\textit{Single-parameter identification}: $E$ recovered to within 0.04\,\% across
all noise levels (0--10\,\%), starting from a 50\,\% initial error.
(2)~\textit{Multi-parameter identification}: simultaneous $E$ and $k_\text{soil}$
identification succeeds with 0.025\,\% and 0.0002\,\% error respectively, even
under 10\,\% measurement noise.
(3)~\textit{Noise robustness under a correctly centred prior}: performance remains stable across noise levels,
with only modest RMSE increase up to 10\,\%; the EMA-adaptive weighting
keeps the data, PDE, boundary, and prior terms on comparable scales as noise changes.
Figure~\ref{fig:inv_noise_robust} visualises this robustness.
(4)~\textit{Training time}: mean 150\,s (2.5 min) for 5{,}000 epochs; inference
$<$0.8\,ms after training.

\textbf{Role and limitation of the Bayesian prior.} The prior is centered at
the \emph{design/as-built specification} value of $E$ (210\,GPa for structural
steel in the reported synthetic cases), which is known from engineering
documentation and represents legitimate domain knowledge when the as-built
stiffness is close to the undamaged design value. In practice, this prior would
be set from material certificates, commissioning tests, or calibrated baseline
inspection data. The wide log-space standard deviation ($\sigma = 1.0$,
covering ${\sim}78$--$570$\,GPa) prevents physically implausible parameter
drift during early training, but the present validation does not prove
robustness when the prior mean is substantially misspecified by degradation.

\subsection{FORM Reliability Assessment}
\label{sec:results_reliability}

\subsubsection{Validation Test Suite}
\label{sec:results_rel_suite}

An eight-case representative suite spanning analytical verification, MPP
optimisation, FORM integration, edge cases, and Monte Carlo cross-validation was
assembled. Table~\ref{tab:form_results} reports the selected cases.

\begin{table*}[htbp]
\caption{FORM reliability validation: selected cases. $\beta_\text{anal}$ =
analytical exact; $\beta_\text{MC}$ = Monte Carlo ($10^5$ samples);
error relative to analytical or MC. Case F3 illustrates a degenerate quadratic
pathology where FORM returns the trivial zero-distance solution rather than a
meaningful reliability index.}
\label{tab:form_results}
\centering
\small
\begin{tabular}{lllccccc}
\toprule
\textbf{Test} & \textbf{Limit State} & \textbf{Dim} &
$\beta_\text{anal}$ & $\beta_\text{FORM}$ & $\beta_\text{MC}$ &
\textbf{Error} & \textbf{Speedup} \\
\midrule
F1 & Linear: $X_1+X_2-10$   & 2D & 4.472 & 4.472 & --   &
$1.8{\times}10^{-15}$ & 24$\times$ \\
F2 & Beam capacity (linear)   & 2D & 3.430 & 3.430 & 3.38 &
$8.9{\times}10^{-16}$ & 26$\times$ \\
F3 & Quadratic pathological case    & 2D & 3.000 & 0.000 & --   &
trivial MPP & -- \\
F4 & Beam bending (nonlin.)  & 4D & --    & 3.824 & 3.65 &
4.7\,\% & 25$\times$ \\
F5 & Triaxial stress          & 4D & --    & 4.010 & 4.26 &
6.0\,\% & 37$\times$ \\
F6 & OWT cantilever root moment  & 3D & --    & 9.900 & 10.00 &
1.0\,\% & $>$40$\times$ \\
F7 & Series system           & 3D & --    & 4.966 & 10.00 &
(MC saturated) & 48$\times$ \\
F8 & High reliability        & 2D & 8.321 & 8.321 & --   &
$<10^{-12}$ & -- \\
\midrule
\multicolumn{8}{@{}p{0.98\linewidth}@{}}{\textbf{Summary:} OWT cantilever case (F6): FORM about 1.0\,ms; MC about 43\,ms ($10^5$ samples);
OWT-case speedup $>$40$\times$; convergence achieved for all well-conditioned test cases in this suite.} \\
\bottomrule
\end{tabular}
\end{table*}

\begin{figure*}[htbp]
\centering
\includegraphics[width=0.48\linewidth]{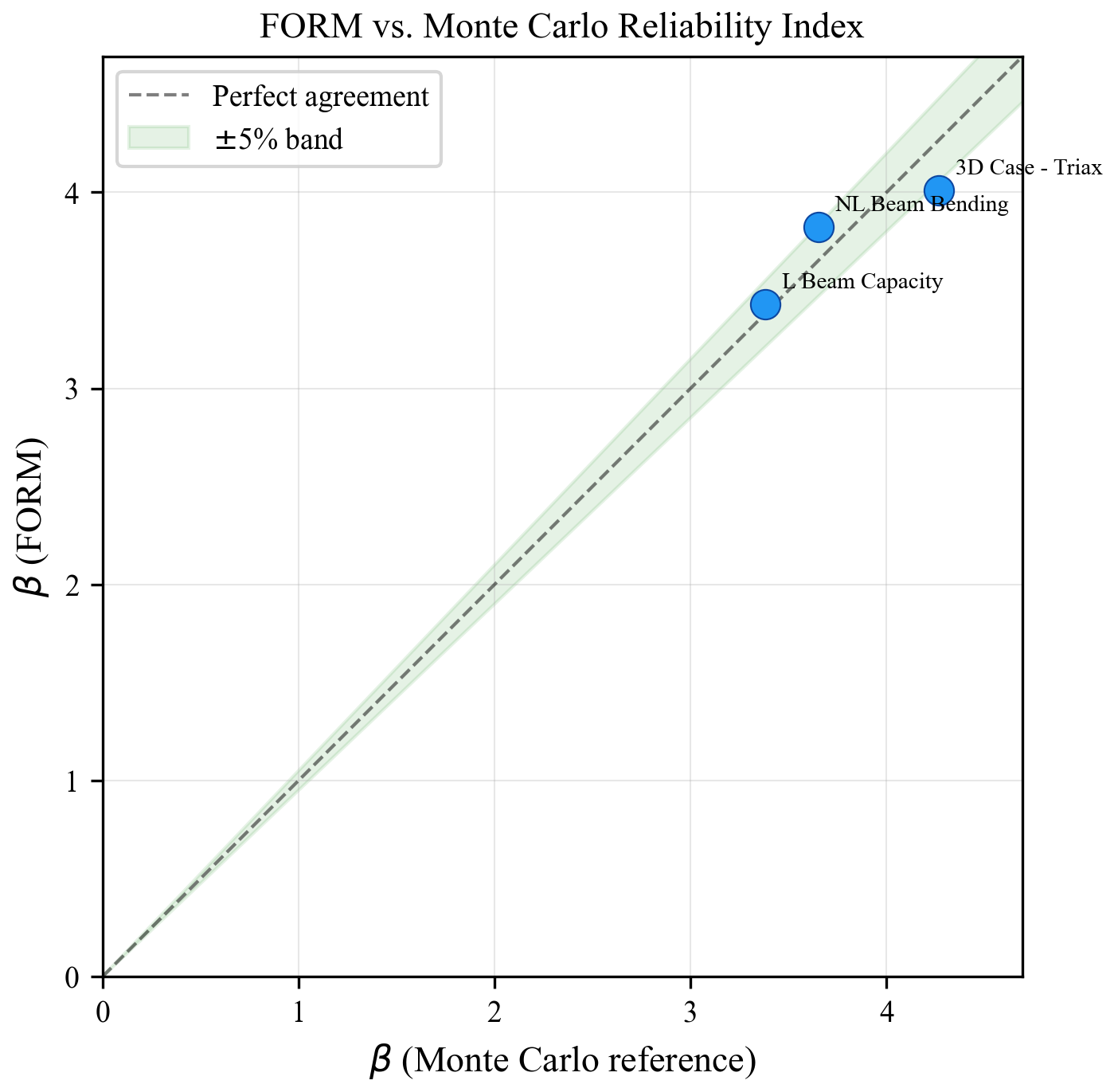}\hfill
\includegraphics[width=0.48\linewidth]{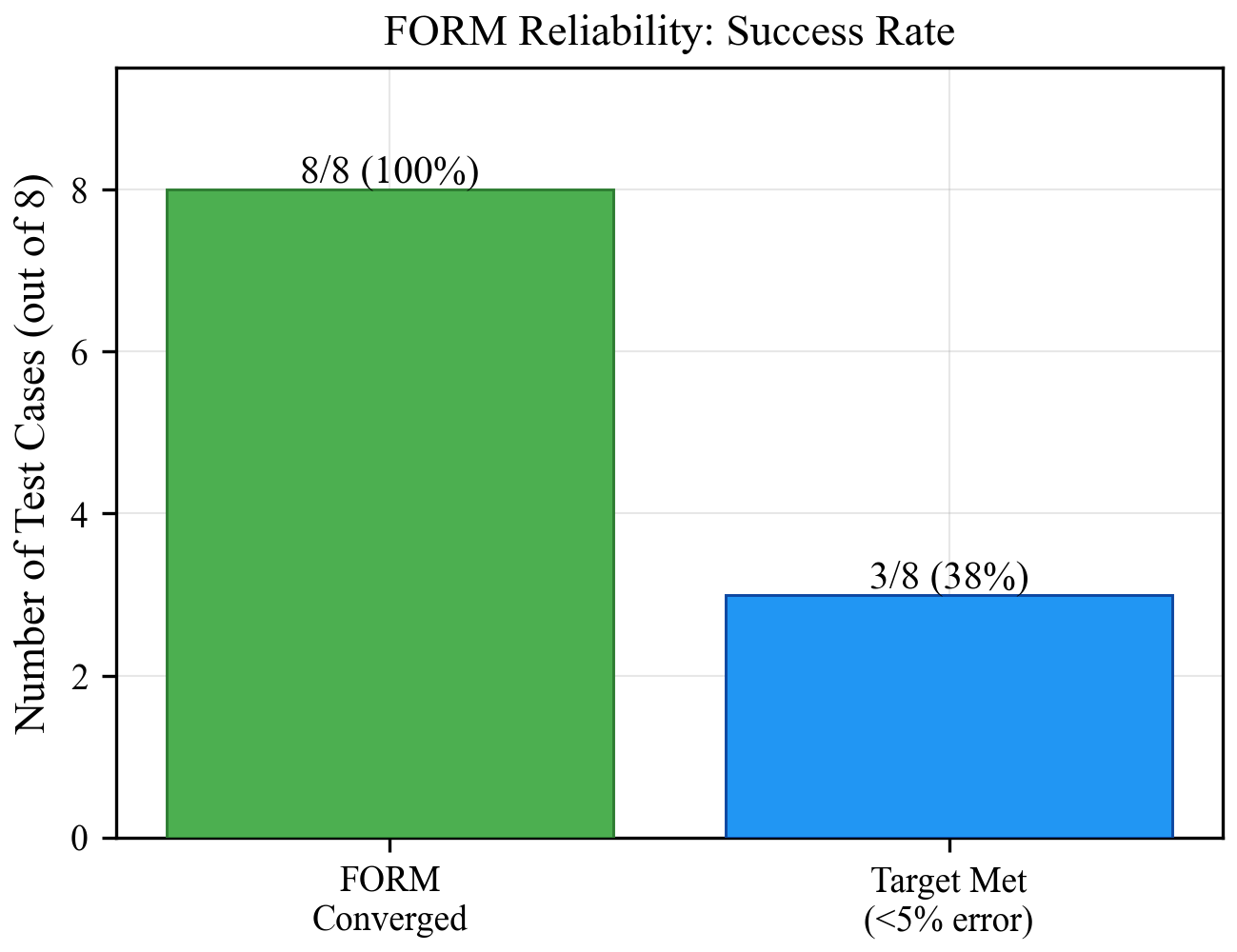}
\caption{\textbf{FORM reliability validation.} Left: FORM vs.\ Monte Carlo
reliability index $\beta$ for test cases with finite MC reference;
machine-precision agreement for linear limit states,
${\leq}6\,\%$ error for moderately nonlinear cases. Right: overall success
rates; 100\,\% FORM convergence for well-conditioned cases in this suite.}
\label{fig:form_perf}
\end{figure*}

\textbf{Linear limit states (F1, F2, F8):} Machine-precision accuracy ($ < 10^{-12}$).
\textbf{Moderately nonlinear (F4--F6):} FORM errors 1.0--6.0\,\%, consistent with
first-order approximation bounds \citep{Wang2022}. Case F3 is a degenerate quadratic
pathology that illustrates a limitation of the gradient-based FORM solver: it
returns the trivial $\beta=0$ solution rather than a physically meaningful
reliability index; 
F7's MC reference saturates at $\beta=10$ (numerical limit).
OWT cantilever-case computation: about 1.0\,ms (FORM, F6) versus about 43\,ms
(Monte Carlo, $10^5$ samples); OWT-case speedup $>$40$\times$ over Monte Carlo.
Figure~\ref{fig:form_perf} summarises FORM validation accuracy and convergence
across the representative test cases.

\subsection{End-to-End Pipeline Performance}
\label{sec:results_e2e}

Table~\ref{tab:e2e} summarises pipeline latency on the reported benchmark
hardware.

\begin{table*}[htbp]
\caption{End-to-end synthetic digital-twin workflow latency. Timings are measured on the
reported GPU/CPU hardware for the synthetic benchmark workflow. FORM timing uses
the reported algebraic benchmark limit states; direct PINN-in-the-loop FORM is
not separately benchmarked. FEA/OpenFAST is shown only as an indicative offline
reference, not as a measured output of this study.}
\label{tab:e2e}
\centering
\small
\begin{tabularx}{\textwidth}{@{}p{0.29\textwidth}p{0.21\textwidth}X@{}}
\toprule
\textbf{Pipeline Stage} & \textbf{Time (ms)} & \textbf{Comment} \\
\midrule
Preprocessing \& normalisation & $<0.1$    & \\
Forward PINN inference         & 0.381 (GPU) / 0.605 (CPU) & 4{,}353 params; all-configuration mean \\
Inverse PINN forward pass (frozen) & $\sim0.75$  & No gradient steps (frozen checkpoint) \\  
FORM $\beta(t)$ computation   & 1.0        & Algebraic OWT limit state; SLSQP, 5--15 evaluations \\
Decision/alert packaging       & $<3.5$     & Conservative allowance; posterior MCMC not benchmarked \\
\midrule
\textbf{Total latency}         & $<7$       & $>$143\,Hz theoretical processing headroom \\
FEA/OpenFAST reference         & --         & Offline high-fidelity analysis; not measured here \\
\textbf{Comparison}            & --         & Surrogate query vs.\ full-model solve is indicative only \\
\bottomrule
\end{tabularx}
\end{table*}

\section{Discussion}
\label{sec:discussion}

\subsection{Forward PINN: Physics as a Data Multiplier}
\label{sec:disc_fwd}

The 10/10 success rate across synthetic test configurations supports the central
premise of PINNs: governing equations provide additional structural constraints
when labelled data are sparse. Compared to wind-turbine component-monitoring
studies surveyed in the literature, where data-driven models commonly use hundreds
of samples \citep{LeonMedina2025}, the present benchmark reaches sub-millimetre
RMSE and $R^2>0.90$ using 50--100 observations. Cross-study accuracy comparisons
remain indicative because metrics such as MAPE are not well conditioned for
clamped-beam displacement fields with near-zero reference values. The 0.381\,ms
mean GPU inference time ($26\times$ below the 10\,ms ceiling) suggests that
edge deployment is computationally plausible, consistent with the edge-computing
architecture of \citep{Ambarita2024}, while embedded-hardware validation is left
for future work.

\subsection{The Gradient Direction Problem: Implications and Generality}
\label{sec:disc_gdp}

The problem is not expected to be specific to the Euler--Bernoulli PDE. By the
gradient-analysis derivation in Appendix~\ref{app:proof}, inverse PINNs where an unknown parameter
$\phi$ multiplies a high-order derivative can exhibit this failure when $\phi$
is initialised far from truth. The 52\,\% initial offset (100 vs.\ 210\,GPa) is
a deliberate large-offset stress test rather than a measured degradation
scenario: corrosion, fatigue, scour, and soil degradation can reduce effective
structural stiffness or make it uncertain over service life
\citep{Augustyn2021,Wang2022}, so inverse solvers should be tested across wide
initialisation ranges. The root cause is structural: displacement data constrains $u(x)$ but not
$u^{(4)}(x)$, trapping~$E$ near its incorrect initial value.

A weak Gaussian prior on $\log E$, centred at the log of the baseline
design/as-built value (available from material certificates or commissioning
tests), breaks this circular dependency when the baseline prior is accurate.
With $\sigma = 1.0$ in log-space ($\sim$78--570\,GPa in the present setting),
the implied log-normal prior on $E$ provides a consistent gradient signal while
still allowing parameter movement around the baseline design value.
Simultaneous $E + k_\text{soil}$ identification succeeds (experiments I3, I4,
Table~\ref{tab:twostage}), suggesting that priors on each nominal parameter can
provide useful regularisation for the coupled multi-parameter case. These
results support the following design principle:
\textit{supplement inverse PINNs with domain-knowledge priors to anchor
parameter search in physically plausible regimes.}

\subsection{FORM Accuracy and SORM Recommendation}
\label{sec:disc_form}

FORM's machine-precision accuracy for linear limit states and ${\leq}6\,\%$ error for
moderately nonlinear benchmark cases covers representative capacity, bending, and
stress checks relevant to OWT support-structure screening. Fatigue and
scour-displacement limit states are compatible with the same FORM interface but
are not separately validated here. The 1.0\,ms OWT cantilever-case evaluation in these benchmark
conditions gives substantial theoretical processing headroom for SCADA-style
monitoring, before field I/O and deployment overheads are included. For strongly
curved or buckling-dominated limit-state surfaces, SORM with principal-curvature
correction should be considered as a future extension.

\subsection{Operational Motivation}
\label{sec:disc_economics}

Using published literature assumptions, a representative 30-turbine offshore
wind farm (8\,MW each) could see approximately \texteuro{}520{,}000 in annual
savings per farm if comparable condition-based maintenance savings were
achieved \citep{LeonMedina2025}. This figure is used only to motivate the
operational value of faster structural state estimation. \textit{The present
synthetic benchmarks do not support any economic-performance, savings-rate, or
payback estimate.}

\subsection{Scope of Claims}
\label{sec:disc_scope}

The contribution should be read as an integration-and-diagnosis benchmark rather
than a field-ready offshore digital twin. The results show that a compact PINN
state estimator, a calibrated-prior inverse identification loop, and FORM
screening can be combined in a one-dimensional synthetic beam workflow with
millisecond-range query time. They do not establish aeroelastic fidelity,
field-transfer accuracy, or robust degradation identification under
substantially misspecified priors.

Table~\ref{tab:scope_assumptions} makes the main assumptions and validation
boundaries explicit for interpreting the reported results.

\begin{table}[!t]
\caption{Assumptions and validation scope for interpreting the benchmark results.}
\label{tab:scope_assumptions}
\centering
\small
\setlength{\tabcolsep}{4pt}
\begin{tabularx}{\columnwidth}{@{}>{\raggedright\arraybackslash}p{0.34\columnwidth} >{\raggedright\arraybackslash}X@{}}
\toprule
\textbf{Scope item} & \textbf{Interpretation and required next step} \\
\midrule
Synthetic validation data & Results are validated against analytical or finite-difference ground truth, not field SCADA or inspection data. Field transfer remains to be demonstrated. \\
One-dimensional support model & The benchmark uses a static Euler--Bernoulli/Winkler representation of the monopile. Deployment-grade claims require 3D dynamic aeroelastic and hydrodynamic validation. \\
Correctly centred prior & Bayesian inverse identification is demonstrated when the weak log-normal prior is centred on the correct design/as-built stiffness. Degraded or misspecified priors require adaptive or hierarchical calibration. \\
Algebraic FORM limit states & Reliability timing and accuracy are shown for analytical benchmark limit states and representative OWT cantilever checks. Direct PINN-in-the-loop FORM, fatigue, scour, and buckling limit states are future work. \\
Operational validation path & OpenFAST and field validation, realistic sensor drift/noise studies, and embedded-hardware tests are required before claiming an operational offshore digital twin. \\
\bottomrule
\end{tabularx}
\end{table}

\subsection{Limitations and Future Work}
\label{sec:disc_limitations}

\textbf{Simulation validation only.} Field validation with real SCADA data is
required. Sensor drift, electromagnetic interference, non-Gaussian noise, scour,
and biofouling must be characterised through instrumented offshore campaigns.
Strain-gauge curvature fusion is a priority extension.

\textbf{Model simplifications.} Static 1D Euler--Bernoulli model; full 3D
dynamic aero-elastic integration via OpenFAST (ServoDyn + AeroDyn) is required
for deployment-grade modelling \citep{NREL2024}. A preliminary time-dependent extension
(\texttt{PINNBeamDynamic}) achieved $R^2 = 0.9531$ on free-vibration data;
full dynamic PINN validation is deferred to future work. Morison forcing
helpers are implemented, but Morison-driven PINN training and end-to-end
field validation are deferred.

\textbf{Inverse problem extensions.} Prior misspecification is the most critical
open question. Preliminary misspecified-prior sweeps in the implementation show
that when the true stiffness is 10--40\,\% below the 210\,GPa design prior, the
current prior-informed training can remain biased toward the design value rather
than recover the degraded state. Deployment therefore requires adaptive or
hierarchical priors calibrated from commissioning data, inspections, or
multi-modal measurements; SORM integration is also needed for buckling limit
states.

\textbf{Online adaptation.} Transfer learning (freeze physics layers, fine-tune
data layers on $\sim$100 new observations, $\approx$11--24\,s) enables
fleet-level amortisation; a full retraining-trigger framework is deferred to
future work.

\section{Conclusions}
\label{sec:conclusion}

This paper presented DigiTurbine as a synthetic reliability-aware PINN benchmark
for offshore wind turbine monopile support-structure monitoring. The study is
best interpreted as an integration-and-diagnosis benchmark rather than a
field-validated offshore digital twin. The following conclusions are drawn.

\textbf{(1) Forward PINN state estimation.}
Across 10 synthetic sparse-data configurations, the forward PINN achieved mean
RMSE $0.135 \pm 0.109$\,mm with all-configuration mean inference of
0.381\,ms on GPU and 0.605\,ms on CPU ($26\times$ / $17\times$ below the
10\,ms target). These results show that physics regularisation can provide
accurate millisecond-range displacement-field reconstruction in the simplified
beam benchmark using 50--100 observations.

\textbf{(2) Inverse PINN failure mode and mitigation.}
Standard simultaneous inverse PINN training failed for the Euler--Bernoulli beam
inverse cases (0/4 pass, mean error 63.9\,\%), and gradient analysis linked this
failure to conflicting data/PDE update directions. Weak Bayesian priors centred
on the correct design/as-built values substantially mitigated the failure in the
reported synthetic tests: 8/8 configurations passed, mean $E$ error was
$\leq 0.02$\,\%, and simultaneous $E{+}k_\text{soil}$ identification remained
successful under 10\,\% measurement noise.

\textbf{(3) FORM reliability screening.}
The FORM layer completed representative limit-state solves in 0.7--2.7\,ms,
including about 1.0\,ms for the OWT cantilever root-moment capacity case
($>40\times$ faster than the reported Monte Carlo reference). Linear limit
states showed machine-precision agreement, while moderately nonlinear cases
remained within ${\leq}6\,\%$ error. The resulting synthetic online benchmark
completed in $<$7\,ms, providing theoretical processing headroom for
SCADA-style monitoring.

Future work should prioritise field or high-fidelity aeroelastic validation,
adaptive priors for degraded or misspecified structures, multi-modal
strain/acceleration fusion, SORM for nonlinear limit states, and fleet-level
transfer learning.

\appendix
\section{Gradient-Analysis Derivation for Standard Inverse PINN Failure}
\label{app:proof}

\textbf{Analytical statement.} \textit{In the standard simultaneous inverse PINN
(Eq.~\ref{eq:joint}), when $\theta$ is randomly initialised and $\phi = E$
is initialised distant from $E^*$, the gradient $\nabla_E\mathcal{L}$ can be
weak or misdirected during training.}

\textbf{Derivation.} For the static case $EI\,u^{(4)} = q$, the PDE loss gradient
with respect to $E$ is:

\begin{equation}
\frac{\partial \mathcal{L}_\text{PDE}}{\partial E}
= \frac{2I}{N_f}\sum_j
  \underbrace{\bigl(E I\hat{u}_\theta^{(4)}(x_j) - q_j\bigr)}_{\text{residual}}
  \cdot
  \underbrace{\hat{u}_\theta^{(4)}(x_j)}_{\text{direction signal}}.
\label{eq:proof1}
\end{equation}

At random initialisation, $\hat{u}_\theta^{(4)}(x_j)$ is near zero (small
positive or negative); for a positive distributed load,
$EI\hat{u}_\theta^{(4)} - q \approx -q < 0$.
The product in Eq.~\eqref{eq:proof1} is therefore dominated by noise with no
consistent sign, providing no reliable signal toward $E^*$.

Crucially, $\partial \mathcal{L}_\text{data}/\partial E = 0$ because $E$
does not appear in the forward prediction $\hat{u} = f_\theta(x)$: the only
gradient signal for~$E$ comes from $\mathcal{L}_\text{PDE}$.
Simultaneously, $\mathcal{L}_\text{data}$ drives the shared network
weights~$\theta$ to fit observed displacements, but the network acquires
fourth derivatives $\hat{u}^{(4)}_\theta$ consistent with the current
(incorrect)~$E$ rather than with~$E^*$.  As shown above,
$\partial\mathcal{L}_\text{PDE}/\partial E$ then evaluates to near-zero at
$E \approx E_\text{current}$, trapping~$E$ and preventing convergence.

Gradient-direction analysis over the full parameter vector
$\Omega = \{\theta,\phi\}$ confirms the conflict: signed cosine similarity
between $\nabla_\Omega \mathcal{L}_\text{data}$ and
$\nabla_\Omega \mathcal{L}_\text{PDE}$ is $-0.21 \pm 0.07$ (mean over 11
runs, multiple seeds), with ${\sim}80\,\%$ of training steps showing opposing
gradient directions in the tested configurations.

\textbf{Expected behaviour without prior.}
In principle, as the data loss $\mathcal{L}_\text{data}$ saturates during
EMA-adaptive joint co-evolution, the balancing should keep the PDE residual
active enough for $\hat{u}_\theta^{(4)}$ to become physically consistent (close
to $q/(E^*I)$).  At that point, the residual $(EI\hat{u}_\theta^{(4)} - q)$
would correctly change sign as $E$ crosses $E^*$, providing a properly directed
gradient.

\emph{Empirical finding.}  For the 4th-order Euler--Bernoulli equation with
displacement-only data and no prior information, this theoretical prediction is \textbf{not} borne out
experimentally.  The network's 4th derivative is dominated by the PDE constraint
at the \emph{current} (incorrect) $E$ value, yielding
$\hat{u}_\theta^{(4)} \approx q/(E_\text{current}\,I)$ rather than
$q/(E^*I)$, which creates a circular dependency that prevents convergence.
However, when a Gaussian prior on $\log E$ centred at
$\log E_\text{design}$ is incorporated into the loss, the prior gradient
provides a consistent signal toward the design-scale stiffness and mitigates this
circular dependency, yielding mean $E$ error $\leq 0.02$\,\% and all
single-parameter noise cases below 0.04\,\% in the tested configurations
(Section~\ref{sec:results_inv}).

\section*{Declaration of Competing Interest}
The author declares no competing financial interests or personal relationships.

\section*{Data Availability}
Synthetic dataset generation scripts and PINN training code available from the
corresponding author on request. OpenFAST was explored for future aeroelastic
validation; input files for the NREL~5\,MW reference configuration are publicly
available at \url{https://www.nrel.gov/wind/nwtc/openfast.html} (no OpenFAST
simulation data were used in the experiments reported here).

\section*{Funding}
This research received no specific external funding.

\section*{Acknowledgements}
The author gratefully acknowledges support from the School of Artificial
Intelligence and Data Science at IIT Jodhpur for access to compute infrastructure
and research facilities. Thanks are also extended to colleagues for productive
discussions on inverse PINN training stability and reliability analysis.

\bibliographystyle{elsarticle-num}
\bibliography{references}

\end{document}